\documentclass[10pt,twoside]{article}
\usepackage{amsmath}
\usepackage[utf8]{inputenc}
\usepackage{url}
\usepackage{dsfont}
\usepackage{graphicx}
\usepackage{yhmath}
\usepackage[table]{xcolor}
\usepackage{mathrsfs} 
\usepackage{amssymb}
\usepackage{url}
\usepackage{makecell}
\usepackage{arydshln}
\usepackage{multirow}
\usepackage{arydshln}
\usepackage{mathtools}
\usepackage{stmaryrd}  
 \usepackage{amsthm,amssymb}
\usepackage{hyperref}

\input epsf
\numberwithin{equation}{section}

\newtheorem{thm}{Theorem}[section]

\newtheorem{Corol}[thm]{Corollary}
\newtheorem{Def}[thm]{Definition\rm}

\newtheorem{rmrk}[thm]{Remark}

\newcommand{\E}{\ensuremath{\mathbb{E}}}
\newcommand{\R}{\ensuremath{\mathbb{R}}}
\newcommand{\Z}{\ensuremath{\mathbb{Z}}}

\newcommand{\N}{\ensuremath{\mathbb{N}}}

\newcommand{\cov}{\ensuremath{\mathrm{Cov}}}

\newcommand{\lip}{\ensuremath{\mathrm{Lip}}}
\definecolor{grisclair}{gray}{0.9}
\font\dsrom=dsrom10 scaled 1200
\def \ind{\textrm{\dsrom{1}}}
\DeclareMathOperator*{\argmin}{argmin}

\begin{document}

\title{\bf Penalized deep neural networks estimator with general loss functions under weak dependence}
 \maketitle \vspace{-1.0cm}

\begin{center}
      William Kengne\footnote{Developed within the ANR BREAKRISK: ANR-17-CE26-0001-01 and the  CY Initiative of Excellence (grant "Investissements d'Avenir" ANR-16-IDEX-0008), Project "EcoDep" PSI-AAP2020-0000000013
   } 
   and 
     Modou Wade\footnote{Supported by the MME-DII center of excellence (ANR-11-LABEX-0023-01)} 
 \end{center}

  \begin{center}
  { \it 
 THEMA, CY Cergy Paris Université, 33 Boulevard du Port, 95011 Cergy-Pontoise Cedex, France\\
  E-mail:   william.kengne@cyu.fr  ; modou.wade@cyu.fr\\
  }
\end{center}

 \pagestyle{myheadings}

\markboth{Penalized deep neural networks estimator with general loss functions}{Kengne and Wade}

\medskip

\textbf{Abstract}:
This paper carries out sparse-penalized deep neural networks predictors for learning weakly dependent processes, with a broad class of loss functions.  
We deal with a general framework that includes, regression estimation, classification, times series prediction, $\cdots$
The $\psi$-weak dependence structure is considered, and for the specific case of bounded observations, $\theta_\infty$-coefficients are also used.  
In this case of $\theta_\infty$-weakly dependent, a non asymptotic generalization bound within the class of deep neural networks predictors is provided.
For learning both $\psi$ and $\theta_\infty$-weakly dependent processes,  oracle inequalities for the excess risk of the sparse-penalized deep neural networks estimators are established.
When the target function is sufficiently smooth, the convergence rate of these excess risk is close to $\mathcal{O}(n^{-1/3})$.
Some simulation results are provided, and application to the forecast of the particulate matter in the Vit\'{o}ria metropolitan area is also considered.

 \medskip
 
{\em Keywords:} Deep neural networks, sparsity, weakly dependence, excess risk, convergence rate.

\medskip

\section{Introduction}
Recent works on deep neural networks (DNNs) have led to significant advances on the theoretical properties of these predictors; see for instance \cite{ohn2019smooth}, \cite{schmidt2019deep}, \cite{bauer2019deep}, \cite{schmidt2020nonparametric}, \cite{tsuji2021estimation}, \cite{kim2021fast} \cite{imaizumi2022advantage} for some results obtained with independent and identical distribution (i.i.d.) observations and \cite{chen2019bbs}, \cite{kohler2020rate},  \cite{kurisu2022adaptive}, \cite{ma2022theoretical}, \cite{kengne2023deep},  \cite{kengne2023excess} in the case of dependent or non-i.i.d. data.  
But, the regularization of DNNs still attracts considerable attention. 
The hidden regularization effects, induced by the learning algorithm, is known as an implicit regularization and is not completely understood nowadays.
Besides, explicit regularizations have been considered in the literature, allowing to increase the accuracy of the DNNs predictors, see, among other papers, \cite{ohn2022nonconvex}, \cite{kurisu2022adaptive}, \cite{kengne2023sparse}, for a sparse-penalized regularization.

\medskip

Consider the observations $ D_n \coloneqq  \{ (X_1, Y_1), \cdots, (X_n, Y_n) \} $ (the training sample) from a stationary and ergodic process $\{Z_t =(X_t, Y_t), t \in \Z \} $, which takes values in $\mathcal{Z}=\mathcal{X}  \times \mathcal{Y} \subset \R^d \times \R $ (with $d \in \N$), where $\mathcal{X}$ is the input space and $\mathcal{Y}$ the output space. Let $  \mathcal{F} \coloneqq \mathcal{F} (\mathcal{X}, \mathcal{Y})$ be the set of measurable functions from $ \mathcal{X}  ~ \text{to} ~\mathcal{Y}$.
In the sequel, we deal with a loss function $\ell: \R \times  \mathcal{Y} \rightarrow [0, \infty) $ and for a predictor $h \in \mathcal{F}$, its risk is given by:
\begin{equation}
 R (h) =  \E_{Z_0} [ \ell (h(X_0), Y_0)].  
\end{equation}
We denote by $h^{*}$, the target predictor (assumed to exist) in  $\mathcal{F} $, satisfying,
\begin{equation}\label{best_pred_F}
R (h^{*}) =  \underset{h \in \mathcal{F}}{\inf} R (h).
\end{equation}
Define for any $h \in \mathcal{F}$, the excess risk,
\begin{equation}
\mathcal{E}_{Z_0}(h) = R (h) - R (h^{*}), ~ \text{with} ~ Z_0 = (X_0, Y_0).
\end{equation} 

\noindent The goal is to build a DNN predictor $\widehat{h}_n $, with an excess risk "close" as possible to zero.

\medskip

We focus on a penalized empirical risk minimization procedure.
%
%
Define the sparse-penalized DNN (SPDNN) predictor by, 
\begin{equation}\label{sparse_DNNs_Estimators}
\widehat{h}_n = \underset{h \in  \mathcal{H}_{\sigma}(L_n, N_n, B_n, F)}{\argmin} \left[ \dfrac{1}{n} \sum_{i=1}^{n} \ell(h(X_i), Y_i) + J_n(h) \right],
\end{equation}
where $ \mathcal{H}_{\sigma}(L_n, N_n, B_n, F) $ is a class of DNNs (see (\ref{DNNs_no_Constraint})) with an activation function $\sigma$ and with suitably chosen architecture parameters $L_n, N_n, B_n, F >0$; and
 $J_ {n} (h) $ denotes a sparse penalty given by, 
\begin{equation}\label{def_Jn}
J_ {n} (h) \coloneqq J_ {\lambda_ {n}, \tau_ {n}} (h) \coloneqq \lambda_ {n} \|\theta(h) \|_ {\text{clip}, \tau_ {n}},
\end{equation}  
for tuning parameters $\lambda_ {n} > 0, \; \tau_ {n} > 0$ and $\theta(h)$ is the network parameter (see (\ref{def_theta_h})), associated to $h$. 
$\| \cdot \|_{\text{clip}, \tau} $ denotes the clipped $ L_1$ norm  with a 
  clipping threshold $ \tau > 0$ see \cite{zhang2010analysis} defined as 
\[ \| \theta \|_{\text{clip}, \tau} = \sum_{j=1}^{p}\left(\frac{|\theta_{j}|}{\tau}\land 
  1\right), \] 
where $\theta= (\theta_1, \cdots, \theta_ p)^T $ is a  $p$-dimensional 
vector and $^T$ denotes the transpose.
In the sequel, we also consider the non penalized predictor, given by,
\begin{equation}\label{NP_DNNs_Estimators}
\widehat{h}_{n, NP} = \underset{h \in  \mathcal{H}_{\sigma}(L, N, B, F, S)}{\argmin} \left[ \dfrac{1}{n} \sum_{i=1}^{n} \ell(h(X_i), Y_i)\right],
\end{equation}
for some network architecture $L, N, B, F, S >0$, where $\mathcal{H}_{\sigma}(L, N, B, F, S)$ denotes a class of sparsity constrained DNNs, defined in (\ref{DNNs_Constraint}).

\medskip

The predictor $\widehat{h}_{n, NP}$, which is based on a sparsity  constraint $S$, enjoys good theoretical properties. 
For regression and classification problems with i.i.d. observations, it can attain optimal convergence rates, see for instance, \cite{schmidt2020nonparametric}, \cite{kim2021fast}. In a general framework with weakly dependent observations, this DNN estimator can achieve a learning rate which is less or close to $\mathcal{O}(n^{-1/3})$, see \cite{kengne2023deep} and Theorem \ref{thm2} below. 
However, the computation of $\widehat{h}_{n, NP}$ is difficult, due to the discrete nature of the sparsity constraint. 
Sparse-penalized regularization have been proposed, among others, to overcome such difficulty.
\cite{ohn2022nonconvex} have performed SPDNN predictor for nonparametric regression and classification of i.i.d. observations. They have established oracle inequalities for the excess risk and derived  convergence rates.
SPDNN estimator for nonparametric time series regression under $\beta$-mixing condition has been proposed by \cite{kurisu2022adaptive}.
A generalization error bound has been provided and they have proved that, this estimator attains the minimax optimal rate up to a poly-logarithmic factor.
\cite{kengne2023sparse} have performed SPDNN predictor for nonparametric regression and classification of $\psi$-weakly dependent processes.
They established oracle inequalities for the excess risk and provided convergences rates, which can asymptotically be close to $\mathcal{O}(n^{-1/2})$. 

\medskip

However, the above works focus on the specific issue of nonparametric regression and/or a particular case of binary classification.
Although \cite{kengne2023sparse} deal with a general class of weakly dependent processes, some restrictions on this paper deserve highlighting: (i) as in \cite{ohn2022nonconvex} and \cite{kurisu2022adaptive}, they focus on nonparametric regression with the $L_2$ loss function; (ii) as in \cite{ohn2022nonconvex} and \cite{kurisu2022adaptive}, the sub-Gaussian assumption is imposed to their model; (iii) their model is quite restrictive, and can not for example, be applied to nonlinear AR-ARCH type processes. 
 
\medskip 

In this contribution, the learning problem by DNNs for a $ \psi$-weakly dependent process $ \left\{ Z_t = \left ( X_t, Y_t \right), ~ t \in \Z \right\} $, with values in $\mathcal{Z}=\mathcal{X}  \times \mathcal{Y} \subset \R^d \times \R $ (with $d \in \N$) is considered.
We deal with a general framework that includes, regression estimation, classification, times series prediction, $\cdots$ and with a broad class of loss functions. A generalization bound under $\theta_\infty$-weakly dependent is derived.
 SPDNN estimator is performed for both $\psi$ and $\theta_\infty$-weakly dependent processes. More precisely, we address the following issues.
\begin{itemize}
\item [(i)] \textbf{Generalization bound for $\theta_\infty$-weakly dependent processes}. We consider the empirical risk minimization (ERM) principle over a class $\mathcal{H}_{\sigma}(L, N, B, F, S)$ (see (\ref{DNNs_Constraint})) for all $L, N, B, F, S >0$.
A non asymptotic bound, with a rate  $\mathcal{O}(n^{-1/4})$ is obtained.
This rate is less than the one obtained in \cite{kengne2023deep} for the non asymptotic generalization bound under $\psi$-weak dependence. 
\item[(ii)] \textbf{Oracle inequalities of the excess risk of the SPDNN predictor under $\psi$ and $\theta_\infty$-weak dependence}. 
Consider the class of sparsity non constrained DNNs predictors $ \mathcal{H}_{\sigma}(L_n, N_n, B_n, F) $ (see (\ref{DNNs_no_Constraint})) for $L_n, N_n, B_n, F >0$.
We provide conditions on $L_n, N_n, B_n, F$ and on $\lambda_n, \tau_n$ (see (\ref{def_Jn})), so that, for the SPDNN predictor $\widehat{h}_n$ (based on a general class of loss functions, see (\ref{sparse_DNNs_Estimators})), oracle inequalities of the excess risk are established for both $\psi$ and $\theta_\infty$-weakly dependent processes.
\item[(iii)] \textbf{Convergence rate of the excess risk of the SPDNN predictor}. When the target function belongs to a class of H\"{o}lder smooth functions and under $\theta_\infty$-weak dependence, we establish a non asymptotic bound with rate $\mathcal{O}(n^{-1/3})$, of the excess risk of the SPDNN predictor $\widehat{h}_n$.
 \item[(iv)] \textbf{Application to affine causal models and for forecasting $PM_{10}$ in the Vit\'oria metropolitan area}.
 We consider a class of affine causal models with exogenous covariates and focus on nonparametric regression.
 For the $L_1$ and $L_2$ loss function, it is shown that, the results in (i), (ii), (iii) apply to such model.
 Application to $PM_{10}$ (particulate matter with a diameter less than $10 \mu m$) in the Vit\'oria metropolitan area, shows that, the SPDNN predictor works a little better than the model proposed in (\cite{diop2022inference}). 
\end{itemize}

\medskip

The rest of the paper is structured as follows. In Section \ref{asump}, we provide some notations, assumptions and recall the definition of the $\psi$ and $\theta_\infty$-weak dependence. The class of DNNs considered are defined in Section \ref{Def_DNNs}.
In Section \ref{Genebound}, we derive a generalization bound for learning $\theta_\infty$-weakly dependent processes by DNNs.
 Section \ref{excess_risk} provides oracle inequalities and a bound of the excess risk of the SPDNNs predictor. 
Applications to nonparametric regression for affine causal models is considered in Section \ref{application} whereas Section \ref{real_data_example} focuses on application to forecasting $PM_{10}$. 
Section \ref{prove} is devoted for the proofs of the main results.

\section{Notations and assumptions}\label{asump}
In the sequel, for two separable Banach spaces $E_1, E_2$ equipped with norms $\| \cdot\|_{E_1} $ and $\| \cdot\|_{E_2} $ respectively, denote by $\mathcal{F}(E_1, E_2) $, the set of measurable functions from $E_1$ to $E_2$.
For any $h \in \mathcal{F} (E_1, E_2) $ and $\epsilon >0$, set $B(h, \epsilon) $ the ball of radius $\epsilon$ of $\mathcal{F} (E_1, E_2) $ centered at $h$, that is,
\[ B (h, \epsilon) = \big\{ f \in \mathcal{F}(E_1, E_2), ~ \| f - h\|_\infty \leq \epsilon \big\},  \]
where $\| \cdot \|_\infty$ denotes the sup-norm defined in (\ref{def_norm_inf}).
Let $\mathcal{H} \subset \mathcal{F}(E_1, E_2) $, the $\epsilon$-covering number $\mathcal{N}(\mathcal{H}, \epsilon) $ of $\mathcal{H} $ represents the minimal 
number of balls of radius $\epsilon$ needed to cover  $\mathcal{H} $; that is,
\[ \mathcal{N}(\mathcal{H}, \epsilon) = \inf\Big\{ m \geq 1 ~: \exists h_1, \cdots, h_m \in \mathcal{H} ~ \text{such that} ~ \mathcal{H} \subset \bigcup_{i=1}^m B(h_i, \epsilon) \Big\}.  \]
For a function $h: E_1 \rightarrow E_2$ and $U \subseteq E_1$, set,
\begin{equation}\label{def_norm_inf}
\| h\|_{\infty} = \sup_{x \in E_1} \| h(x) \|_{E_2}, ~ \| h\|_{\infty,U} = \sup_{x \in U} \| h(x) \|_{E_2},
\end{equation}  
 and
\begin{equation}\label{def_Lip}
\lip_\alpha (h) \coloneqq \underset{x_1, x_2 \in E_1, ~ x_1\ne x_2}{\sup} \dfrac{ \|h(x_1) - h(x_2)\|_{E_2}}{\| x_1- x_2 \|^{\alpha}_{E_1}} ~ \text{for any}  ~ \alpha \in [0, 1]. 
\end{equation}
 For any $\mathcal{K}_{\ell} > 0$ and $\alpha \in [0, 1]$, $\Lambda_{\alpha,\mathcal{K}_{\ell}} (E_1, E_2) $ (or simply $\Lambda_{\alpha, \mathcal{K}_{\ell}} (E_1) $ when $E_2 \subseteq \R$) is the set of functions  $h: E_1^u  \rightarrow E_2$ for some $u \in \N$, satisfies  $\|h\|_{\infty} < \infty$ and  $\lip_\alpha(h) \leq \mathcal{K}_{\ell}$.
When $\alpha = 1$, we set  $\lip_1 (h) = \lip(h) $ and $ \Lambda_{1} (E_1) =\Lambda_{1, 1}(E_1, \R) $. 

\medskip

Let us define the weak dependence structure, see \cite{doukhan1999new} and \cite{dedecker2007weak}. Let $E$ be a separable Banach space.

\begin{Def}\label{Def}
$ An\; E- valued \; process \;(Z_t)_{t \in \Z}\; is\; said\; to\; be\; (\Lambda_1(E), \psi,\epsilon)-weakly\; dependent\; $  $ if\; there\; exists\; a\;\\ function\; $ $\psi:[0, \infty)^2 \times \N^2 \to [0, \infty) \;and\; $ $a\; sequence\; \epsilon=(\epsilon(r))_{r \in \N}\; decreasing\; to\; zero\; at\; infinity\; such\; that,\; for\;\\ any\; g_1, \;g_2 \in \Lambda_1 (E)\; with\; g_1: E^u \rightarrow \R, \;g_2: E^v \rightarrow \R\;, ~ (u, v \in \N)\; and\; for\; any\;  u-tuple\; (s_1, \cdots, s_u)\; and\; any\; v-tuple\; (t_1, \cdots, t_v)\; with\; s_1 \leq \cdots \leq s_u \leq s_u + r \leq t_1 \leq \cdots \leq t_v, \; the\; following\; inequality\; is\; fulfilled$:
\[ \vert \cov (g_{1}(Z_{s_1}, \cdots, Z_{s_u}),  g_{2}(Z_{t_1}, \cdots, Z_{t_v})) \vert \leq \psi(\lip(g_1),\lip(g_2), u, v) \epsilon(r). \]
\end{Def}
For example, following choices of $\psi$ leads to some well-known weak dependence conditions. 
\begin{itemize}
\item $\psi \left(\lip(g_1),\lip(g_2), u, v \right) = v \lip(g_2) $: the $\theta$-weak dependence, then denote $\epsilon(r) = \theta(r) $;
\item $\psi \left(\lip(g_1),\lip(g_2),u,v \right)= u \lip(g_1) + v \lip(g_2)$: the $\eta$-weak dependence, then denote $\epsilon(r) = 
 \eta(r) $;
\item $\psi \left(\lip(g_1),\lip(g_2), u, v \right)= u v \lip(g_1) \cdot \lip(g_2) $: the $\kappa$- weak dependence, then denote $\epsilon(r) = \kappa(r) $;
\item $\psi \left(\lip(g_1), \lip(g_2), u, v \right) = u \lip(g_1) + v \lip(g_2) + u v \lip(g_1) \cdot \lip(g_2) $: the $\lambda$-weak dependence, then denote $\epsilon(r) = \lambda(r) $. 
\end{itemize}
In the sequel, for each of the four choices of $\psi$ above, set respectively,
\begin{equation}\label{Psi}
\Psi (u, v) = 2 v, ~ \Psi (u, v) = u + v, ~ \Psi (u, v) = u v, ~ \Psi(u, v) = (u + v + u v)/2.
\end{equation}
\medskip
 
We consider the  process $\{ Z_t=(X_t, Y_t), t \in \Z \}$ with  values in $ \mathcal{Z}=   \mathcal{X} \times \mathcal{Y} \subset \R^d \times \R$, the loss function 
$\ell:\R \times \mathcal{Y} \rightarrow [0, \infty)$, the class of DNNs $\mathcal{H}_{\sigma}(L_n, N_n, B_n, F) $ ($L_n, N_n, B_n, F>0$) with the activation function $\sigma: \R \rightarrow \R $, and set of following  assumptions.
\begin{itemize}
\item[\textbf{(A1)}]: There exists  a constant $C_{\sigma} > 0$ such that the activation function $\sigma \in \Lambda_{1, C_{\sigma}}(\R) $.
\item[\textbf{(A2)}]: There exists $\mathcal{K}_{\ell} > 0$ such that, the loss function $\ell \in \Lambda_{1, \mathcal{K}_{\ell}}(\R \times \mathcal{Y})$ and $M_n={\sup_{h\in \mathcal{H}_{\sigma}(L_n, N_n, B_n, F)}}{\sup_{z \in \mathcal{Z}}|\ell(h, z) |} < \infty$. 
 \end{itemize}
 Under \textbf{(A2)}, we get that,
 \begin{equation}\label{def_Gn}
 G_n := \sup_{h_1, h_2 \in  \mathcal{H}_{\sigma}(L_n, N_n, B_n, F), h_1 \neq h_2 } \sup_{ z\in \mathcal{Z}  } \frac{ |\ell(h_1,z) - \ell(h_2,z) | }{ \|h_1 - h_2 \|_\infty } < \infty .
 \end{equation}
Let us set the weak dependence assumption.
\begin{itemize}
\item[\textbf{(A3)}]: Let  $\Psi: [0,  \infty)^2 \times \N^2 \rightarrow [0, \infty)$ be one of the choices in (\ref{Psi}). The process $\left\{ Z_t = (X_t, Y_t), {t \in \Z} \right\}$  is stationary ergodic and   $(\Lambda_1 (\mathcal{Z}), \psi, \epsilon)$-weakly dependent  such that, there exists $L_1, ~ L_2, ~  \mu \ge 0$ satisfying 
\begin{equation}
\sum_{j \ge 0} \left ( j  + 1 \right)^{k} \epsilon_j \leq L_1 L_2^k  (k!)^{\mu} \quad \text{for all} ~ k \ge 0.
\end{equation}
\end{itemize}
In the sequel, we also use the $\theta_{\infty, k} (1)$-coefficients of \cite{dedecker2007weak}, which is a version of $\gamma$-mixing of \cite{rio1999theorie}, see also \cite{rio2000inegalites}. 
\begin{Def}\label{def2}
When $\{ Z_t=(X_t, Y_t), t \in \Z \}$ is a bounded stationary and ergodic process, define for any $k > 0\;$, the $\theta_{\infty, k} (1)$-weak dependence  coefficients by,
\begin{equation}\label{dependencecoef}
\theta_{\infty, k} (1) \coloneqq \underset{h \in \Lambda_{1}(\mathcal{Z}), 0 < j_1 < \cdots < j_k}{\sup}  \Big \|  | \E[h(Z_{j_1}, \cdots, Z_{j_k}) \vert Z_t, t \leq 0] -  \E[h(Z_{j_1}, \cdots, Z_{j_k})] | \Big\|_{\infty},
\end{equation}
where  $ \| V \|_{\infty}$ refers to the essential supremum of  the random variable $\| V \|$, for a  random vector $V $,  and $ \Lambda_{1}(\mathcal{Z})$ is  defined  in Section \ref{asump}.
\end{Def}

\medskip

\noindent \textbf{(A4)} The process $ \{Z_t = (X_t, Y_t ),  t \in \Z \}$ is bounded, stationary and ergodic, with the $ \theta_{\infty, k} (1)$-weakly dependent coefficients satisfying for all $k>0$, $\theta_{\infty, k} (1) \leq \mathcal{C}$, for some $\mathcal{C} >0$.

\section{ Deep Neural Networks }\label{Def_DNNs}
Let us recall that, a DNN architecture $ (L,   \textbf{p}) $ stands for a positive integer $L$ called the number of hidden layers or depth and a width vector $\mathbf{p} = (p_0,p_1,\cdots,p_{L+1}) \in \N^{L+2}$.
So, a DNN with network architecture $(L, \mathbf{p})$ is any function of the form,
\begin{equation}\label{h_equ1}
h: \R^{p_0} \rightarrow \R^{p_{L+1}}, \;   x\mapsto h(x) = A_{L+1} \circ \sigma_{L} \circ A_{L} \circ \sigma_{L-1} \circ \cdots \circ \sigma_1 \circ A_1 (x),
\end{equation} 
where $ A_j: \R^ {p_ {j -1}} \rightarrow \R^ {p_j} $ is a linear affine map, defined by $A_j (x) \coloneqq W_j x + \textbf{b}_j$,  for given  $p_ {j - 1}\times p_j$  weight matrix   $ W_j$   and a shift vector $ \textbf{b}_j \in \R^ {p_j} $, and $\sigma_j: \R^{p_j} \rightarrow \R^ {p_j} $ is a nonlinear element-wise activation map, defined for all $z=(z_1,\cdots,z_{p_j})^T$ by $\sigma_j (z) = (\sigma(z_1), \cdots, \sigma(z_{p_j}))^{T} $, and $T$ denotes the transpose.   
For a DNN of the form (\ref{h_equ1}), denote by $\theta(h)$ the vector of its parameters, that is,
\begin{equation} \label{def_theta_h}
\theta(h) \coloneqq \left(vec(W_1)^ {T}, \textbf{b}^{T}_{1}, \cdots,  vec(W_{L + 1})^{T} , \textbf{b}^ {T}_{L+1} \right)^{T}, 
 \end{equation} 
 where $ vec(W)$ denotes the vector obtained by concatenating the column vectors of the matrix $W$. 
 In the sequel, we deal with an activation function $ \sigma: \R \rightarrow \R$ and let us denote by $\mathcal{H}_{\sigma, p_0, p_ {L+1}} $, the class of DNNs predictors with $p_0$-dimensional input and $p_ {L+1} $-dimensional output.
$p_0 = d$ and $p_ {L + 1} = 1$ in the framework considered here.
For a DNN $h$ as in (\ref{h_equ1}), denote by depth($h$) and width($h$) its depth and width respectively; that is, depth($h$)$=L$ and width($h$) = $\underset{1\leq j \leq L} {\max} p_j $. For any positive constants $L, N,   B$ and $F$, set
\[ \mathcal{H}_{\sigma}(L, N, B) \coloneqq \big \{h\in \mathcal{H}_{\sigma, q, 1}: \text{depth}(h)\leq L, \text{width}(h)\leq N, \|\theta(h)\|_{\infty} \leq B \big\},  \]
 and
\begin{equation}\label{DNNs_no_Constraint}
\mathcal{H}_{\sigma}(L, N, B, F) \coloneqq \big\{ h: h\in H_{\sigma}(L, N, B), \| h \|_{\infty, \mathcal{X}} \leq F \big\}. 
\end{equation}
Define also the class of sparsity constrained DNNs with sparsity level $S > 0$ by 
\begin{equation}\label{DNNs_Constraint}
\mathcal{H}_{\sigma}(L, N, B, F, S) \coloneqq \left\{h\in \mathcal{H}_{\sigma}(L, N, B, F) \; :  \; \| \theta(h) \|_0 \leq S 
  \right\},
\end{equation}
where $\| x \|_0 = \sum_{i=1}^p \ind(x_i \neq 0), ~ \| x\| = \underset{1 \leq i  \leq p}{\max} |x_i |$ for all $x=(x_1,\ldots,x_p)^{T} \in \R^p$ ($p \in \N$).

%
%

\section{Generalization bound under $\theta_\infty$-weak dependence}\label{Genebound}

This section focuses on the generalization bound of the ERM algorithm for $\theta_\infty$-weakly dependent processes over a class of DNNs predictors $\mathcal{H}_{\sigma} (L, N, B, F, S) $, for all $L, N, B, F, S>0$. 
The following theorem provides a uniform exponential concentration inequalities over the class $\mathcal{H}_{\sigma} (L, N, B, F, S) $, between the risk and its empirical version.
\begin{thm}\label{thm1}
Assume that \textbf{(A1)}, \textbf{(A2)} (with $L_n=L, N_n=N, B_n=B$ and $M_n=M$) and \textbf{(A4)} hold. For all $n \in \N, \varepsilon > 0$ and $\nu_0 \in (0, 1) $ we have, 
\begin{align}\label{equa2}
\nonumber  & P \Bigg\{  \underset{h \in  \mathcal{H}_{\sigma} (L, N, B, F, S)}{\sup} \Bigg [R (h) - \widehat{R}_n (h) \Bigg] > \varepsilon \Bigg\} 
\\
& \leq  \mathcal{N} \Bigg ( \mathcal{H}_{\sigma} (L, N, B, F, S), \dfrac{\varepsilon}{4 G} \Bigg) \exp\Bigg ( -n^{\nu_0} \varepsilon +  \dfrac{ n^{2 \nu_0 - 1} (M + \theta_{\infty, n} (1))^2}{2}\Bigg),
\end{align}
 where $ \theta_{\infty, n} (1)$ is defined in (\ref{dependencecoef}) and $G=G_n$ given in (\ref{def_Gn}) with  $L_n=L, N_n=N, B_n=B$.
\end{thm}

\medskip

Consider the predictor $\widehat{h}_{n, NP}$ (see (\ref{NP_DNNs_Estimators})) obtained by minimizing the empirical risk. 
Let $h_{\mathcal{H}_{\sigma}}$ be a target neural network (assumed to exist) and defined by
\begin{equation}\label{def_h_H_sigma}
h_{\mathcal{H}_{\sigma}} = \underset{h \in  \mathcal{H}_{\sigma}(L, N, B, F, S)}{\argmin} R(h).
\end{equation} 
The following theorem provides a generalization bound.
\begin{thm}\label{thm2}
 Assume that the conditions of Theorem \ref{thm1} hold. Let $ \eta,  \nu_0 \in (0, 1)$. Assume that  
 \[ n >  \max \Bigg ( n_0 (L, S, M, \nu_0, \theta_{\infty, n} (1)),   \Bigg (  \dfrac{1}{ M} \left ( C_1 -   2 L (S + 1)  \log (2 M) -  \log (\eta) \right)_{+} \Bigg)^{1/2 \nu_0} \Bigg),
 \]
 where
 \[ C_1 = 2 L (S + 1) \log \left( 4 G C_{\sigma} L (N + 1)(B \lor  1) \right), ~  n_0 (L, S, M, \nu_0, \theta_{\infty, n}(1)) \in \N ~ \text{defined at (\ref{equan0})}.          \]
 %


\textbf{(i)} With probability at least $ 1 - \eta$, we have  
\begin{equation}\label{equaERM1}
R \left( \widehat{h}_{n,NP} \right) - \widehat{R}_n \left( \widehat{h}_{n,NP}  \right) \leq \varepsilon \left ( n, \eta,  \nu_0 \right),
\end{equation}
where $ \varepsilon \left ( n, \eta, \nu_0 \right)  <  \dfrac{2 M}{n^{\nu_0/2}}$ and $\widehat{h}_{n,NP}$ is defined in (\ref{NP_DNNs_Estimators}).

\medskip

\textbf{(ii)}  With probability at least $ 1 - 2 \eta$,  we have
\begin{equation}\label{excesrisk}
 R \left (\widehat{h}_{n,NP} \right) -   R \left ( h_{\mathcal{H}_{\sigma}} \right) \leq \varepsilon \left ( n, \eta,  \nu_0 \right) +\varepsilon' \left ( n, \eta, \nu_0 \right),
\end{equation}
where $ \varepsilon' \left (n, \eta, \nu_0 \right) =   \dfrac{(M + \theta_{\infty, n} (1))^2}{2 n^{1 - \nu_0}} + \dfrac{\log (1/\eta)}{n^{\nu_0}} $.
\end{thm}
\begin{rmrk}
\noindent The learning rates of the non asymptotic bounds (\ref{equaERM1}) and (\ref{excesrisk}) are of order $\mathcal{O} (n^{-1/4})$. 
\end{rmrk}

\section{Oracle inequalities of the excess risk of the SPDNN predictor}\label{excess_risk}

In this section, we provided some oracle inequalities of the excess risk of the SPDNN under $\psi$ and $\theta_\infty$-weak dependence conditions.
A bound of this excess risk is derived when the target predictor $h^*$ (see (\ref{best_pred_F})) belongs to a class of H\"older functions.
  In the sequel, we use the notations, $a \lor b = \max(a,b)$, 
$ a_n \lesssim b_n$ or  $ b_n \gtrsim  a_n$  if there exists a positive constant $C> 0$ such that $ a_n \leq  C b_n$ for all $ n \in \N$, $a_n \asymp b_n$ if $ a_n \lesssim b_n$ and $ a_n \gtrsim b_n$.
%
 %
%
%
%
In Theorem \ref{thm4} below, denote $ C_{1, n} = 4M^{2}_n\Psi(1, 1)L_{1} \; \text{and} \;  C_{2, n}= 2M_n L_{2}\max(\frac{2^{3 + \mu}}{\Psi(1, 1)}, 1)$, where $\Psi$  is one of the functions in $(\ref{Psi})$, under assumption \textbf{(A2)} and \textbf{(A3)}. 
\begin{thm}\label{thm4}
 Assume that (\textbf{A1})-(\textbf{A3}) hold.  Let $ L_n \lesssim \log n, N_n \lesssim n^{\nu_1}, 1 \leq B_n \lesssim n^{\nu_2}, F > 0$, for some $ \nu_1 > 0, \nu_2 > 0$. Then, the SPDNN estimator given by,
\begin{equation}\label{EMRpen}
\widehat{h}_n = \underset{ h \in \mathcal{H}_{\sigma} (L_n, N_n, B_n, F)}{\argmin} \Bigg[ \dfrac{1}{n} \sum_{i = 1}^{n} \ell \left( h(X_i), Y_i \right) +  \lambda_n \| \theta (h) \|_{clip, \tau_n}  \Bigg],
\end{equation}
with $ 
\lambda_n \asymp (\log n)^{\nu_3}/ n^{\nu_4}, \tau_n \leq  \dfrac{ \rho_n}{4 \mathcal{K}_{\ell} (L_n + 1)((N_n + 1) B_n)^{L_n + 1}} $ where  $ \rho_n \coloneqq \dfrac{\left( C_{1, n}/(2 C_{2, n}^{1/(\mu + 2)})\right)^{(\mu + 2)/(2 \mu + 3)}}{n^{(\mu + 1) / (2\mu + 3)}}$, satisfies
\begin{align}
\nonumber \mathcal{E}_{Z_0} (\widehat{h}_n) 
& \leq 2  \Bigg\{ \underset{h \in \mathcal{H}_{\sigma}(L_n, N_n, B_n, F)} {\inf} \Bigg[ \mathcal{E}_{Z_0} (h)  + J_{\lambda_n, \tau_n} (h) \Bigg]  \Bigg\} \lor   \dfrac{C_{\mu} \left( C_{1, n}/(2 C_{2, n}^{1/(\mu + 2)})\right)^{(\mu + 2)/(2 \mu + 3)}}{n^{(\mu + 1) / (2\mu + 3)}},
\end{align}
for all $n \ge N$, for some $N \in \N, C_{\mu} >0$ depending on $ \mu >0$, $\mathcal{K}_{\ell} > 0$, for $\nu_3, \nu_4 >0$ such that $ \nu_1  + \nu_2  + \nu_4 < \dfrac{1}{\mu + 2}$. 
\end{thm}

The next theorem deals with $\theta_\infty$-weak dependence conditions.
\begin{thm}\label{thm_3}
Assume that (\textbf{A1})-(\textbf{A2}) and (\textbf{A4}) are satisfied. Let $ L_n \lesssim \log n, N_n \lesssim n^{\nu_1}, 1 \leq B_n \lesssim n^{\nu_2}, F > 0$, for some $ \nu_1 > 0, ~  \nu_2 > 0$. Then, the SPDNN estimator given by,
\begin{equation}\label{EMRpen}
\widehat{h}_n = \underset{ h \in \mathcal{H}_{\sigma} (L_n, N_n, B_n, F)}{\argmin} \Bigg[ \dfrac{1}{n} \sum_{i = 1}^{n} \ell \left( h(X_i), Y_i \right) +  \lambda_n \| \theta (h) \|_{clip, \tau_n}  \Bigg],
\end{equation}
with $ \lambda_n \asymp (\log n)^{\nu_3}/ n^{\nu_4}, ~ \tau_n \leq  \dfrac{ \rho_n}{4 \mathcal{K}_{\ell} (L_n + 1)((N_n + 1) B_n)^{L_n + 1}} $ where  $ \rho_n = 1 /n^{2 \nu_6}, ~ \mathcal{K}_{\ell} > 0$, satisfies
\begin{equation}
\E \Big [\mathcal{E}_{Z_0} (\widehat{h}_n) \Big ] \lesssim 2  \Bigg\{ \underset{h \in \mathcal{H}_{\sigma} (L_n, N_n, B_n, F)} {\inf} \Bigg[ \mathcal{E}_{Z_0} (h)  + J_{\lambda_n, \tau_n} (h) \Bigg]  \Bigg\} \lor \dfrac{C}{n^{2 \nu_6}},
\end{equation}
for all $n \ge N$, for some $  N \in \N, ~ C > 0, ~ \nu_3 > 0, ~ \nu_4 > 0,  ~ \nu_5 \in (0,1)$ such that $\nu_4  + 2 \nu_6 + \nu_1 + \nu_2 < \nu_5$ and  $\nu_6 < \dfrac{ 1 - \nu_5}{2}$.
\end{thm}

\medskip

Theorem \ref{thm4} and \ref{thm_3} above can be used to derive a bound of the excess risk of the SPDNN predictor as shown in Corollary \ref{corol1}).
 Let us consider a class of H\"older smooth functions.
For $ U \subset \R^d, \beta = (\beta_1, \cdots, \beta_d)^T \in \N^d, \quad x = (x_1, \cdots, x_d)^T \in U$, set
\[ |\beta|= \sum_{i =1}^{d} \beta_i ~ \text{and} ~ \partial^{\beta} = \dfrac{\partial^{|\beta|}}{\partial^{\beta_1} x_1, \cdots, \partial^{\beta_d} x_d}.  \]
For any $ s > 0$, the H\"older space $ \mathcal{C}^s (U)$ is a set of functions $ h: U \rightarrow \R$ such that, for any $ \beta \in \N^d$ with $ |\beta| \leq [s], ~ \| \partial^{\beta} h \|_{\infty} < \infty$ and for any $ \beta \in \N^d$ with $ |\beta| = [s], ~ \lip_{s - [s]} (\partial^{\beta} h) < \infty$ (where $[x]$ denotes the integer part of $x$). This space is equipped with the norm
\[ \| h\|_{\mathcal{C}^{s} (U)} =  \underset{0 \leq |\beta| \leq [s]}{\sum} \| \partial^{\beta} h \|_{\infty} +  \underset{|\beta| = [s]}{\sum} \lip_{s - [s]} (\partial^{\beta} h).       \]
 For any $ s > 0, ~  U \subset \R^d $ and $ \mathcal{K} > 0$, set 
\[\mathcal{C}^{s, \mathcal{K}}(U) = \{h \in  \mathcal{C}^{s} (U), ~ \|h\|_{ \mathcal{C}^{s} (U)} \leq \mathcal{K}\}. \]

\begin{Corol}\label{corol1}
Assume that (\textbf{A1})-(\textbf{A2}), (\textbf{A4}) hold and that $h^* \in \mathcal{C}^{s, \mathcal{K}}(\mathcal{X})$ for some $\mathcal{K} > 0$, where $h^*$ is defined in (\ref{best_pred_F}). Let $ L_n, N_n, B_n, F > 0, \lambda_n, \tau_n, \rho_n$ given as in Theorem (\ref{thm_3}), with $ \nu_1 = \dfrac{d \nu_4}{s(1 + d/s)}, ~ \nu_2 = \dfrac{4 d \nu_4}{(s + 1)(1 + d/s)} $ and $\nu_5 \in (0, 1)$. Consider the DNNs class $ \mathcal{H}_{\sigma}(L_n, N_n, B_n, F) $ where the activation function $\sigma$ is  either piecewise linear or locally quadratic and fixes a segment $I \subseteq [0,1]$ (see \cite{kengne2023deep}).
  Then, the SPDNN estimator defined in (\ref{EMRpen}) satisfies,
\begin{equation}\label{excess_risk_bound}
\E \Big [\mathcal{E}_{Z_0} (\widehat{h}_n) \Big ] \lesssim \dfrac{  2 \nu_4(\log n)^{\nu_3 + 1}}{(1 + d/s) n^{\nu_4/(1 + d/s)}} \lor \dfrac{C}{n^{2 \nu_6}},
\end{equation}
for all $n \ge N$  for some  $  N \in \N, ~ C > 0, ~ \nu_3 > 0, ~ \nu_4 > 0,  ~ \nu_4  + 2 \nu_6 + \nu_1 + \nu_2 < \nu_5$ and $\nu_6 < \dfrac{ 1 - \nu_5}{2}$.
\end{Corol}
\noindent From (\ref{excess_risk_bound}), one can see that, if the target predictor $h^*$ is sufficiently smooth such that $s\gg d$, by choosing $\nu_1, \nu_2$ close to 0, $\nu_5$ close to 2/3, we get that, the convergence rate of the excess risk of the SPDNN predictor is close to $\mathcal{O}(n^{-1/3})$.


\section{ Application to nonparametric regression}\label{application}
%
%
%
In this section, we perform nonparametric regression by DNNs for a large class of autoregressive models. 
\subsection{Affine causal models}
Let $(\mathcal{X}_t)_{t \in \Z}$ be a process of covariates with values in $ \R^{d_x} ~ (d_x \in \N) $.
Consider the  class of affine causal models with exogenous covariates (see (\cite{diop2022inference})) given by
\begin{equation}\label{models_af}
Y_t = f (Y_{t - 1}, \cdots, Y_{t - p}; \mathcal{X}_{t - 1}) + \mathcal{M} (Y_{t - 1}, \cdots, Y_{t - p}; \mathcal{X}_{t - 1}) \xi_t, 
\end{equation}
where $\mathcal{M}, f: \R^{p } \times \R^{d_x} \rightarrow \R, ~ (p \in \N) $ are two measurable functions, and $(\xi_t)_{t \in \Z}$ is a sequence of i.i.d. centered random variable satisfying $\E [\xi_0^r] < \infty$ for some $r \ge 2$ and $\E [\xi_0^2] = 1$.
The following Lipschitz-type conditions on the functions $f, \mathcal{M} ~ \text{or} ~ \mathcal{M}^2$ are set by \cite{diop2022inference} in order to study the stability properties of the models (\ref{models_af}).
Let us denote by $0$ the null vector of any vector space. For $ \Psi = f  ~ \text{or} ~ \mathcal{M} $, let us consider the assumptions:

\medskip

\noindent \textbf{Assumption A($\Psi$)}: $| \Psi (0; 0)| < \infty $ and there exists a sequence of non-negative real numbers $ (\alpha_{k, Y} (\Psi))_{1\leq k \leq p} $ and $ \alpha_{\mathcal{X}} (\Psi) >0$, satisfying $ \sum_{k = 1}^{p} \alpha_{k, Y} (\Psi) < \infty, ~ \alpha_{\mathcal{X}} (\Psi) < \infty$; such that, for any $(y, u), ~ (y', u') \in \R^p \times \R^{d_x}$,
\[ |\Psi(y; u) - \Psi(y'; u') | \leq \sum_{k = 1}^{p} \alpha_{k, Y} (\Psi) |y_k - y'_k| + \alpha_{\mathcal{X}} (\Psi) |u - u'|.  \]
We set the following assumption on the function $H = \mathcal{M}^2$ for  ARCH-X type models.

\medskip

\noindent \textbf{Assumption A(H)}: Assume that $f = 0, ~ |\mathcal{M}(0; 0)| < \infty$ and there exists a sequence of non-negative real numbers $ (\alpha_{k, Y}(H))_{1 \leq k \leq p} $ and $ \alpha_{\mathcal{X}} (H) >0$ satisfying $ \sum_{k = 1}^{p} \alpha_{k, Y}(H) < \infty, ~ \alpha_{\mathcal{X}}(H) < \infty$; such that, for any $(y, u), ~ (y', u') \in \R^p \times \R^{d_x}$,
\[ | H(y, u) - H(y', u') | \leq \sum_{k = 1}^{p} \alpha_{k, Y}(H) |y^2_k - y'^2_k| + \alpha_{\mathcal{X}}(H) |u - u'|.  \]
The convention if \textbf{A}$(\mathcal{M})$ holds, then $ \alpha_{k, Y}(H) = \alpha_{\mathcal{X}}(H) = 0$ for all $1 \leq k \leq p$ and if \textbf{A}$ (H) $ holds, then $ \alpha_{k, Y} (\mathcal{M}) = \alpha_{\mathcal{X}}(\mathcal{M}) = 0$ for all $1 \leq k \leq p$ is made for the following. An autoregressive-type structure is imposed to the covariates: 
\begin{equation}\label{def_covar}
\mathcal{X}_t = g(\mathcal{X}_{t - 1}, \mathcal{X}_{t - 2}, \cdots; \eta_t),
\end{equation}
where $ (\eta_t)_{t \in \Z}$ is a sequence of i.i.d. random vectors with values in $ \R^{d_{\eta}}$ ($d_{\eta} \in \N$) and $ g (u; \eta)$ is a measurable function with values in $\R^{d_x}$, satisfying
\begin{equation}\label{covar_condi}
\E [\| g (0; \eta_0) \|^r] < \infty ~ \text{and} ~ \| g (u; \eta_0) - g (u'; \eta_0) \|_r \leq \sum_{k = 1}^{\infty} \alpha_k (g) \|u_k - u'_k \| ~ \text{for all} ~  u, u'  \in \left(\R^{d_x} \right)^{\infty},
\end{equation}
for some $ r \ge 1$, and a non-negative sequence $(\alpha_k(g))_{k \ge 1} $ satisfying $ \sum_{k = 1}^{\infty} \alpha_{k}(g) < 1$; where $ \|U \|_r \coloneqq (\E \|U \|^r)^{1/r}$ for any random vectors $U$. Let us consider the model (\ref{models_af}) with the conditions (\ref{def_covar}) and (\ref{covar_condi}) on the covariates where $ (\eta_t)_{t \in \Z}$ in (\ref{def_covar}) is such that, $ (\eta_t, \xi_t)_{t \in \Z}$ is a sequence of i.i.d. random vectors, and we assume that for some $ r \ge 1$,
\begin{equation}\label{cond_lip}
\sum_{k = 1}^{p} \max \Big\{ \alpha_{k} (g), \alpha_{k, Y} (f) + \| \xi_0 \|_r \alpha_{k, Y} (\mathcal{M}) +  \| \xi_0 \|^2_r \alpha_{k, Y} (H) \Big\}
+
\sum_{k = p+1}^{\infty} \max \{ \alpha_{k} (g) \} < 1.
\end{equation}
Under the assumptions $ A(f), ~ A(\mathcal{M}), ~ A (H)$, and (\ref{cond_lip}), there exists a $\tau$-weakly dependent stationary, ergodic and non adaptive  solution  $ (Y_t, \mathcal{X}_t)_{t \in \Z}$ of (\ref{models_af}) satisfying $ \| (Y_0, \mathcal{X}_0)\|_r < \infty$ (see (\cite{diop2022inference})). This solution is $ \theta$-weakly dependent with coefficients $ \theta(j)$ bounded as $ j \rightarrow \infty$ by,
\begin{equation}
\theta (j) \leq \tau (j) = \mathcal{O}\Bigg( \underset{1 \leq \iota \leq j}{\inf} \Bigg\{ \alpha^{j/\iota} + \underset{k \ge \iota + 1}{\sum} \alpha_k \Bigg\} \Bigg),
\end{equation}
where $ \alpha_k =\max \Big\{ \alpha_{k} (g), \alpha_{k, Y} (f) + \| \xi_0 \|_r \alpha_{k, Y} (\mathcal{M}) +  \| \xi_0 \|^2_r \alpha_{k, Y} (H) \Big\}$ for $1 \leq k \leq p$, $\alpha_k = \alpha_{k} (g)$ for $k \geq p + 1$ and $\alpha = \sum_{k \ge 1} \alpha_k$.
This solution also satisfies the $\theta_\infty$-weak dependence condition in the assumption (\textbf{A4}), see \cite{alquier2012model}.

\medskip

We focus on the nonparametric estimation of the function $f$ in (\ref{models_af}) by SPDNNs. 
Set,
\[X_t = (Y_{t - 1}, \cdots, Y_{t - p}; \mathcal{X}_{t - 1}).  \]
If we use the quadratic loss function, then, a target predictor is given for all $x \in \R^{p } \times \R^{d_x}$ by
\begin{equation}
h^{*} (x) = \E[ Y_0 | X_0 =x] = f(x).
\end{equation}

\noindent If we deal with the $L_1$ loss function, then, a target predictor is given by
\begin{equation}
h^{*} (x) = \text{med} \left( Y_0 | X_0 = x\right),
\end{equation}
where $\text{med}(V)$ denotes the median of $V$, for any random variable $V$.
We consider innovations $ (\xi_t)_{t \in \Z}$ in (\ref{models_af}) and $ (\eta_t)_{t \in \Z}$ in (\ref{def_covar}) generated from a standardized uniform distribution $ \mathcal{U} [-a; a] $ and $ \mathcal{U} [-a'; a']$ respectively  for some $a,a' >0$ and assume that $\mathcal{M}(x)>0$ for all $x \in \R^{p } \times \R^{d_x}$.
In this case, one can get that $ \text{med} \left( Y_0 | X_0 = x\right) = f (x) $.
So, for the nonparametric estimation of $f$ in (\ref{models_af}) by SPDNNs, under the conditions (\ref{def_covar}) and (\ref{cond_lip}), the results of Theorem \ref{thm4} and \ref{thm_3} can be applied. In addition,  $\mathcal{Z}=\mathcal{X} \times \mathcal{Y} \subset \R^d \times \R^{d_x}$ can be chosen to be bound, if $f$ belongs to a H\"older space $ \mathcal{C}^s (\mathcal{Z})$ for some $s>0$, then, Corollary \ref{corol1} can be applied.

\subsection{Example of nonlinear ARX-ARCH models}\label{nonlinear_model}
%
%
%
%
We consider a nonlinear ARX-ARCH(1) models defined by
\begin{align}\label{auto_regr_model}
\left\{
\begin{array}{ll}
Y_t & = f (Y_{t - 1}, Y_{t - 2}; \mathcal{X}_{t - 1}) + \varepsilon_t \\
\varepsilon_t & = \xi_t \sqrt{\phi_0 + \phi_1\varepsilon_{t - 1}^2},
\end{array}
\right.
\end{align}
where $ (\xi_t)_{t \in \Z}$ represents innovations generated from a standardized uniform distribution $ \mathcal{U}[-2; 2] $ and $(\mathcal{X}_t)_{t \in \Z}$ is an AR(1) process defined by
\begin{equation}\label{equa_covariate}
\mathcal{X}_t = \alpha_0 + \alpha_1 \mathcal{X}_{t- 1} + \eta_t, \quad  \text{with} ~ \alpha_0 \in \R, ~ |\alpha_1| < 1,
\end{equation}
generated from innovations $ (\eta_t)_{t \in \Z}$ that follows a standardized uniform distribution $\mathcal{U}[-2; 2]$, such that, $ (\eta_t, \xi_t)_{t \in \Z}$ is a sequence of i.i.d. random vectors.
One can write model (\ref{auto_regr_model}) in the form (\ref{models_af}) as:
\begin{equation}\label{particular_auto_regr_model}
Y_t = f(Y_{t - 1}, Y_{t - 2}; \mathcal{X}_{t - 1}) + \xi_t \sqrt{\phi_0 + \phi_1 \left(Y_{t - 1} - f(Y_{t - 2}, Y_{t - 3}; \mathcal{X}_{t - 2}) \right)^2}.
\end{equation}
Set,
\[X_t = (Y_{t - 1}, Y_{t - 2}; \mathcal{X}_{t - 1}).           \]
Let us set a Lipschitz condition on $f$. 
Assume that there exist $ \alpha_{1, Y}(f), \alpha_{2, Y}(f), \alpha_{\mathcal{X}}(f) >0$ such that, for any $ x = (y_1, y_2; u), ~ x' = (y'_1, y'_2; u') \in \R^2 \times \R^{d_x}$: 
\begin{align}\label{equa_lip_f}
|f(x) - f(x')| \leq \alpha_{1, Y}(f) |y_1 - y'_1| + \alpha_{2, Y}(f) |y_2 - y'_2| + \alpha_{\mathcal{X}}(f)|u - u'|.
\end{align}
Set,
\[ \mathcal{M}(X_t) = \sqrt{\phi_0 + \phi_1 \left(Y_{t - 1} - f(X_{t - 1}) \right)^2}. \]
We have for all $ x = (y_1, y_2; u), ~ x' = (y'_1, y'_2; u') \in \R^2 \times \R^{d_x}$,
\begin{align}\label{equa_lip_M}
\nonumber \left| \mathcal{M}(x) -  \mathcal{M}(x') \right| & = \left|\sqrt{\phi_0 + \phi_1 (y_1- f(x))^2} - \sqrt{\phi_0 + \phi_1 (y'_1- f(x'))^2} \right|
\\
\nonumber & = \dfrac{|\phi_1| \left|(y_1- f(x))^2 - (y'_1- f(x'))^2 \right|}{\sqrt{\phi_0 + \phi_1 (y_1 - f(x))^2} + \sqrt{\phi_0 + \phi_1 (y'_1- f(x'))^2}} 
\\
\nonumber & \leq \dfrac{|\phi_1| \left|(y_1 - f(x)) - (y'_1 - f(x'))\right|}{\sqrt{\phi_1}}
\\
\nonumber & \leq \sqrt{\phi_1} \big(|y_1 - y'_1|+  \alpha_{1, Y}(f)|y_1 - y'_1| + \alpha_{2, Y}(f) |y_2 - y'_2| + \alpha_{\mathcal{X}}(f)|u - u'| \big)
\\
& \leq \alpha_{1, Y}(\mathcal{M}) |y_1 - y'_1| + \alpha_{2, Y}(\mathcal{M}) |y_2 - y'_2| + \alpha_{\mathcal{X}}(\mathcal{M})|u - u'|,
\end{align}
with $\alpha_{1, Y}(\mathcal{M}) =\sqrt{\phi_1} \left(1 + \alpha_{1, Y}(f) \right), ~ \alpha_{2, Y}(\mathcal{M}) = \sqrt{\phi_1} \alpha_{2, Y}(f), ~ \alpha_{ \mathcal{X}}(\mathcal{M}) = \sqrt{\phi_1} \alpha_{ \mathcal{X}}(f)$. We will focus on (\ref{cond_lip}) with $ r = 2$, and we assume that the following condition is fulfilled,
\begin{equation}\label{lip_cond_M}
\max \Bigg\{ \alpha_1, \alpha_{1, Y}(f) + \alpha_{1, Y}(\mathcal{M}) \Bigg\} + \alpha_{2, Y}(f) + \alpha_{2, Y}(\mathcal{M}) < 1.
\end{equation}
Thus, under the conditions (\ref{equa_covariate}), (\ref{equa_lip_f}), (\ref{equa_lip_M}) and (\ref{lip_cond_M}), there exists a solution $(Y_t, \mathcal{X}_t)_{t \in \Z} $ of (\ref{auto_regr_model}), which satisfying assumptions (\textbf{A3}) and (\textbf{A4}) above.

\medskip
Let $(Y_1, \mathcal{X}_1), \cdots, (Y_n, \mathcal{X}_n) $ be a trajectory of the process $ (Y_t, \mathcal{X}_t)_{t \in \Z}$ satisfying (\ref{auto_regr_model}).
We aim to predict $ Y_{n + 1}$ based on this sample. 
We perform the learning theory above with the SPDNN predictors with the input variable $X_t = (Y_{t - 1}, Y_{t - 2}; \mathcal{X}_{t - 1}) \in \mathcal{X} \subset \R^3$  and, the output space $\mathcal{Y} \subset \R$. We consider the model (\ref{auto_regr_model}) with $\phi_0 = 0.25, ~ \phi_1 = 0.1$ in the following case of $f$:
\[
\begin{array}{ll}
\text{DGP1}: f(Y_{t - 1}, Y_{t - 2}, \mathcal{X}_{t - 1})  = -0.75 + 0.1 \max(Y_{t - 1}, 0) - 0.2 \min(Y_{t - 1}, 0) + 0.15 Y_{t - 2} + 0.4 \sqrt{1 + 0.5 \mathcal{X}_{t - 1}^2}; & 
\\
\text{DGP2}: f(Y_{t - 1}, \mathcal{X}_{t - 1}) = 0.4 + (0.2 - 0.15 e^{-Y_{t - 1}^2}) Y_{t - 1} - \dfrac{0.5}{1 + \left|\mathcal{X}_{t - 1} \right|}, &  
\end{array} 
\]
where $ (\mathcal{X}_t)_{t \in \Z}$ is an AR(1) process satisfying (\ref{equa_covariate})  with $ \alpha_0 = \alpha_1 = 0.5$.
DGP1 is a Threshold autoregressive model with nonlinear covariate, and DGP2 is an exponential autoregression model with nonlinear covariate.

\medskip

\noindent In the sequel, we deal with the $ L_2$ and $L_1$ loss functions, whose, a target function are respectively,
\begin{equation}
h^{*}(x) = \E[Y_0| X_0 = x] = f(x), 
\end{equation}
and,
\begin{equation}
h^{*}(x) = \text{med} (Y_0 | X_0 = x) = f(x),
\end{equation}
for all $x \in \mathcal{X}$ which can be chosen to be a compact subset of $\R^3$, and med represents the median.

\medskip

For each of these DGPs, we carry out a network architecture of 2 hidden layers with 100 hidden nodes for each layer. We use the ReLU activation function in the hidden layers and the linear activation function in the output layer. We trained the network weights thanks to the package keras in the R software, by using the algorithm Adam (see (\cite{kingma2014adam})) with learning rate $10^{-3}$ and the minibatch size of $32$. We stopped the training when the mean square error (MSE) does not improve within $30$ epochs. 

\medskip

We generated a trajectory $ \left( (Y_1, \mathcal{X}_1), (Y_2, \mathcal{X}_2), \cdots, (Y_n, \mathcal{X}_n) \right)$ from the true DGP for $ n = 250, ~ 500, ~ \text{and} ~ 1000$.
The predictor $\widehat{h}_n$ from (\ref{sparse_DNNs_Estimators}) is derived with the tuning parameters are of the form $ \lambda_n = 10^{-i} \log(n)/n, ~ \text{and} ~ \tau_n = 10^{-j}/\log(n) $, for $i, j = 0, 1, \cdots, 10$ and are calibrated by minimizing the mean squared error (MSE) and the mean absolute error (MAE) on a validation data set 
$ \left( (Y'_1, \mathcal{X}'_1), (Y'_2, \mathcal{X}'_2), \cdots, (Y'_n, \mathcal{X}'_n) \right)$. The excess risk of the SPDNN predictor $\widehat{h}_n$ is then calculated based on a new trajectory $ \left( (Y''_1, \mathcal{X}''_1), (Y''_2, \mathcal{X}''_2), \cdots, (Y''_m, \mathcal{X}''_m) \right) $ with $ m = 10^4$.  

\medskip

\begin{figure}[h!]
\begin{center}
\includegraphics[height= 19.96cm, width=18.98cm]{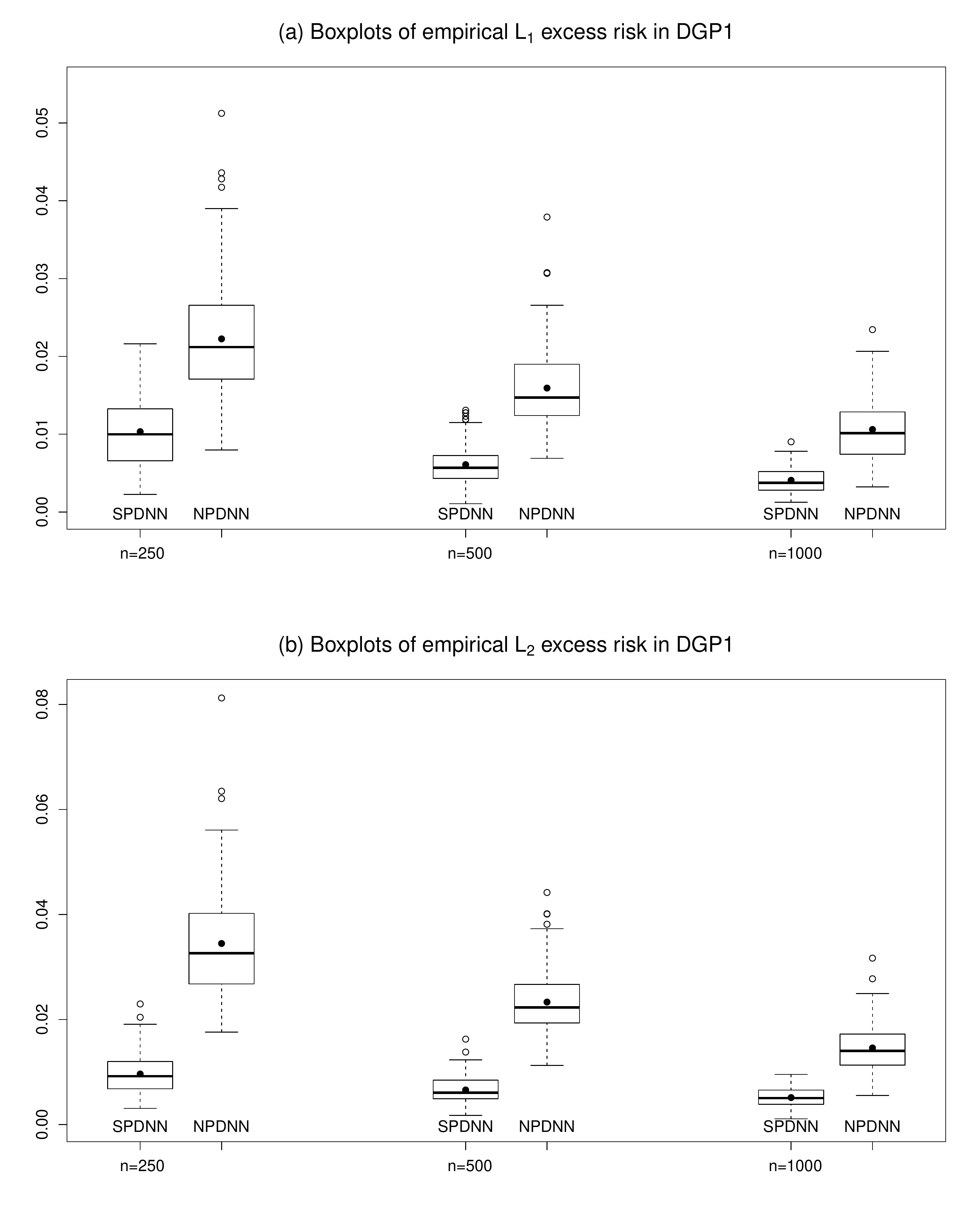}
\end{center}
\vspace{-.7cm}
\caption{\it Boxplots of empirical $L_1$ excess risk (a) and empirical $L_2$ excess risk (b)  of the SPDNN and NPDNN predictors with $n=250, 500$ and 1000 in DGP1.}
\label{Graphe_DGP_1}
\end{figure}

\begin{figure}[h!]
\begin{center}
\includegraphics[height= 19.96cm, width=18.98cm]{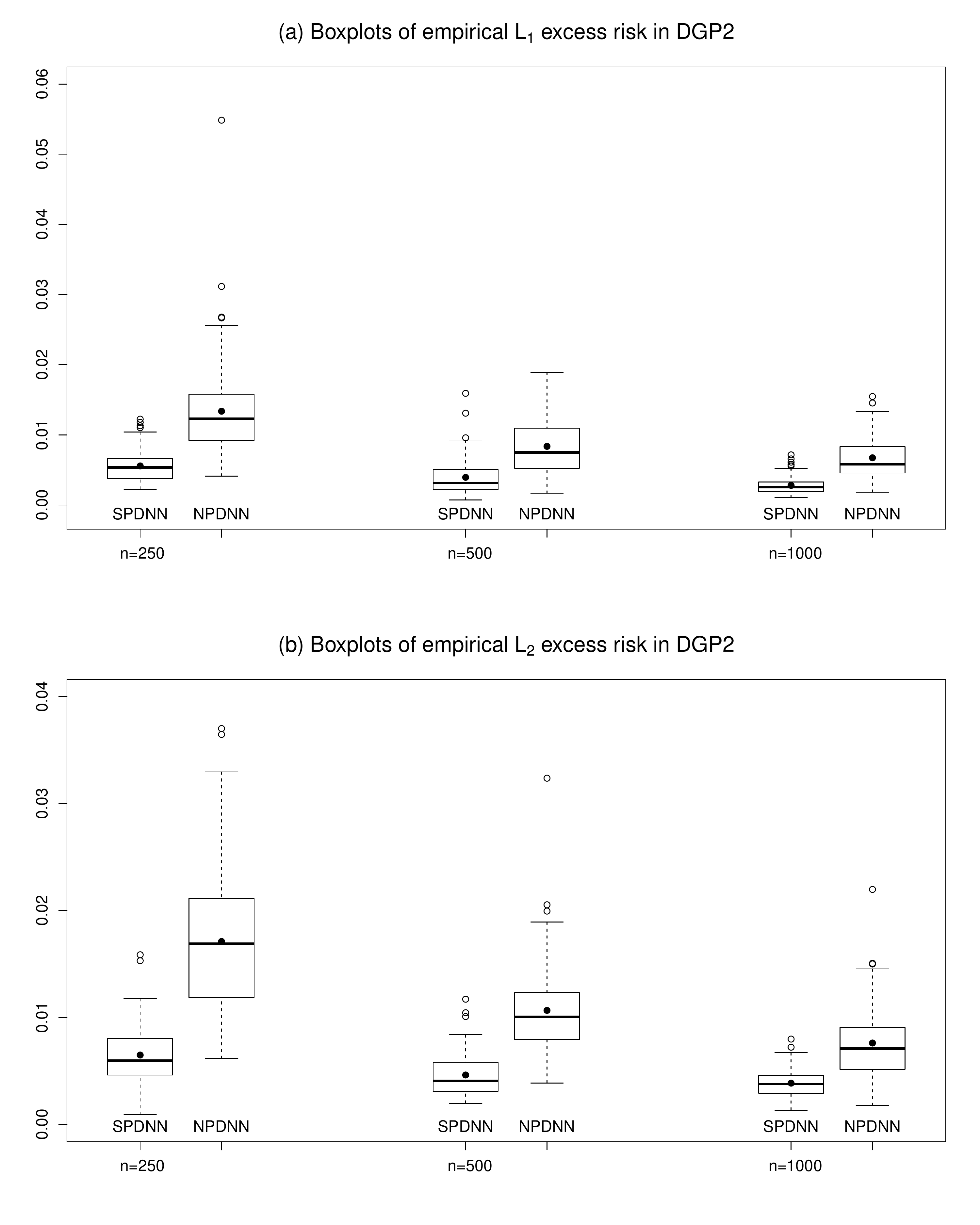}
\end{center}
\vspace{-.7cm}
\caption{\it Boxplots of empirical $L_1$ excess risk (a) and empirical $L_2$ excess risk (b)  of the SPDNN and NPDNN predictors with $n=250, 500$ and 1000 in DGP2.}
\label{Graphe_DGP_2}
\end{figure}

Figure (\ref{Graphe_DGP_1}) and (\ref{Graphe_DGP_2}) display the boxplots of the empirical $L_1$ excess risk (a) and the empirical $L_2$ excess risk (b) of the SPDNN and non penalized DNN (NPDNN, obtained from (\ref{sparse_DNNs_Estimators}) with $\lambda_n=0$) predictors over 100 replications. For both DGP1 and DGP2, one can see that, the empirical excess risk overall decreases as the sample size increases for $L_1$ and $L_2$ loss; which is in accordance with the theoretical results.
Also, the results displays by the SPDNN estimator are overall more accurate than those of the NPDNN estimator. This finding shows that, the accuracy of the prediction can be improved by the SPDNN predictor with respect to the NPDNN.
%

\section{Real data example}\label{real_data_example}
\begin{figure}[h!]
\begin{center}
\includegraphics[height= 6.96cm, width=18.98cm]{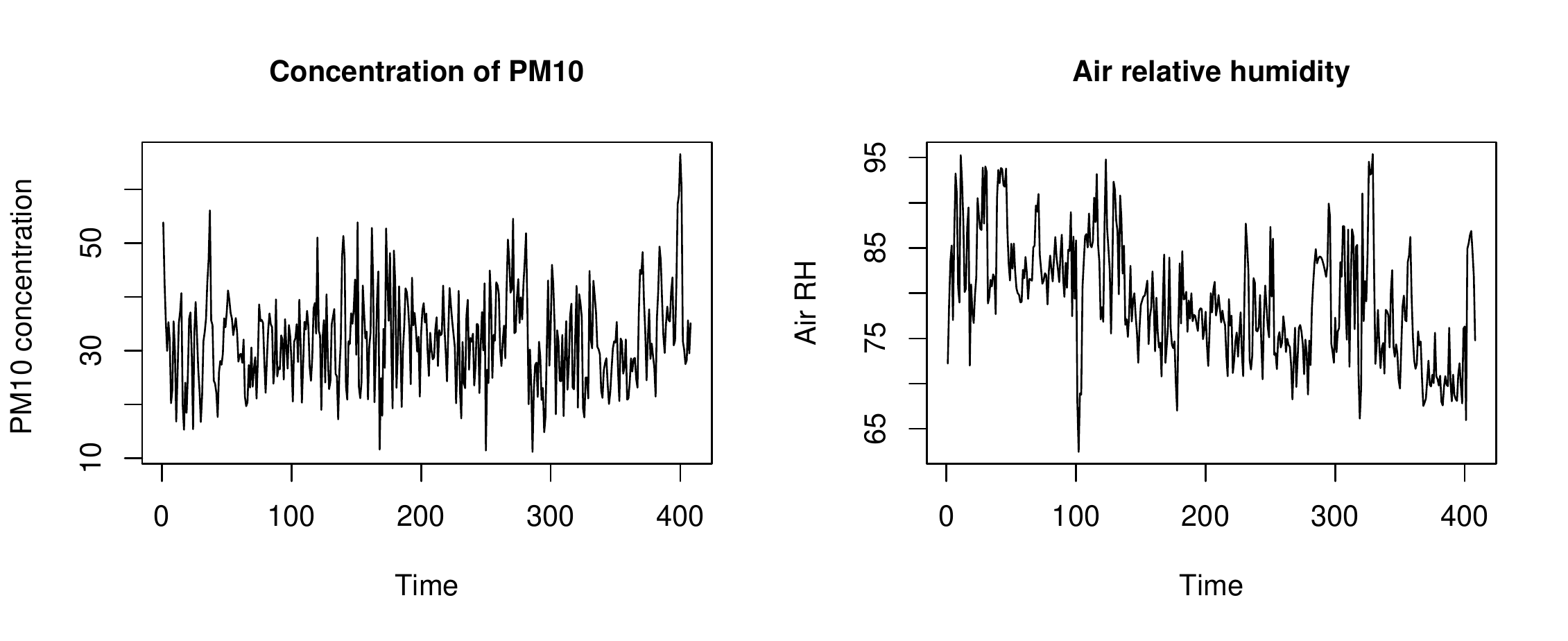}
\end{center}
\vspace{-.7cm}
\caption{\it The daily concentrations of the PM10 and the Air relative humidity from January 21st, 2005 to March 04th 2006, in the Vit\'oria metropolitan area.}
\label{Graphe_PM10_RH}
\end{figure}

%
%
%
We consider the daily concentration of the $PM_{10}$ (particulate matter with a diameter less than $10 \mu m$) in the Vit\'oria metropolitan area; from January $21\text{st}, 2005 ~ \text{to March} ~ 04 \text{th}, ~ 2006$. 
Figure \ref{Graphe_PM10_RH} shows the concentration of the $PM_{10}$  and the air relative humidity (RH) during the considered period. These data  are obtained from the State Environment and Water Resources Institute, and were collected at eight monitoring stations. We focus on the $408$ observations that are a part 
of a large dataset (available at \url{https://rss.onlinelibrary.wiley.com/pb-assets/hub-assets/rss/Datasets/RSSC\%2067.2/C1239deSouza-1531120585220.zip}) which were performed by \cite{souza2018generalized} to evaluate the association between respiratory disease and air pollution concentrations.

\medskip

We  consider the nonparametric regression task and we aim to predict the output variable $Y_t = PM_ {10, t}$ for a given  input $X_t = (PM_{10, t - 1}, RH_{t - 1})$. \cite{diop2022inference} have performed a double autoregressive (DAR) model with exogenous covariates on these data and have obtained,
\begin{equation}\label{concentration_PM10}
PM_ {10, t} = 37.946 + 0.330 PM_ {10, t - 1} - 0.210 RH_{t - 1} + \xi_t \sqrt{32.108 + 0.023 PM_{10, t - 1}^2},
\end{equation}
where $(\xi_t)_{t \in Z}$ is a centered i.i.d. random variable, and $(RH_{t})_{t \in \Z}$ is the air relative humidity covariate.
We consider the $L_2$ loss function, so, according to the DAR model  (\ref{concentration_PM10}), a target predictor is,
\begin{equation}\label{Pred_PM10}
 PM^{DAR}_ {10, t} = \E \left[PM_ {10, t}| (PM_ {10, t - 1}, RH_{t - 1}) \right] = 37.946 + 0.330 PM_ {10, t - 1} - 0.210 RH_{t - 1}.
\end{equation}
 For the estimation task, we perform a network architecture as in the previous section (2 hidden layers, 100 hidden nodes for each layer and the ReLU activation function) and with input variable $X_t = ( PM_{10, t - 1}, RH_{t - 1})$.

\begin{figure}[h!]
\begin{center}
\includegraphics[height= 19.96cm, width=17.98cm]{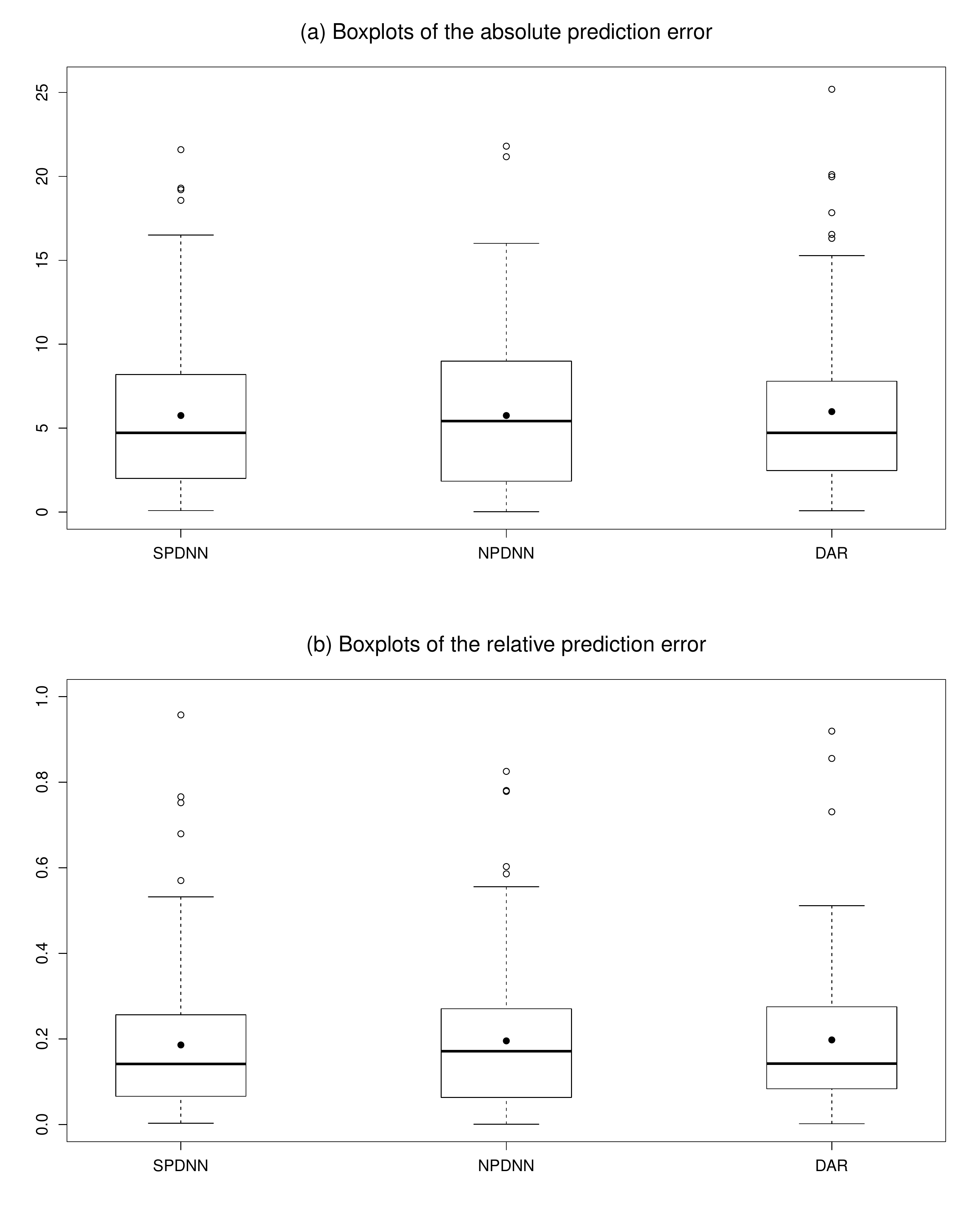}
\end{center}
\vspace{-.7cm}
\caption{\it Boxplots of the absolute prediction error (a) and the relative prediction error (b) of the SPDNN, NPDNN and DAR predictors.}
\label{res_real_data}
\end{figure}

\medskip
We consider predictions based on the SPDNN, NPDNN estimator and the DAR predictor (\ref{Pred_PM10}).
For the SPDNN and NPDNN predictors, the networks are trained with the first 308 observations and the rest is used as test data.
For each of the predictors SPDNN, NPDNN and DAR, the mean absolute prediction error and the mean relative prediction error given by,
\begin{align*}
\text{mean abs.pred.error} & =  \dfrac{1}{100} \sum_{t = 1}^{100} \left|PM_{10, t} - \widehat{PM}_{10, t} \right| 
\\
\text{mean relat.pred.error} & = \dfrac{1}{100} \sum_{t = 1}^{100}\dfrac{\left|PM_{10, t} - \widehat{PM}_{10, t} \right|}{PM_{10, t}},
\end{align*}
is computed, where $\widehat{PM}_{10, t}$ represents the prediction at time $t$.
Figure \ref{res_real_data} (a) and (b) displays the boxplots of the absolute and the relative prediction error.
Table \ref{tab1} displays the values of the  mean absolute prediction error and the mean relative prediction error for each of the predictors SPDNN, NPDNN and DAR.
One can see from Table \ref{tab1} that, the mean relative prediction error as well as the mean absolute prediction error of the SPDNN predictor is slightly better than those of the NPDNN and DAR predictors.
This is despite that, the predictions obtained here based on the SPDNN estimators are out-of-sample predictions, whereas those obtained from the DAR model are in-sample predictions. 
\begin{table}
\centering
\begin{tabular}{|c|c|c|c|}
\hline
predictor & Mean relative prediction error (\%)  & Mean absolute prediction error   \\
\hline
SPDNN&5.75&18.50\\
\hline
NPDNN&6.03&19.55\\
\hline
DAR&6.01&19.79 \\
\hline
\end{tabular}
\caption{Some performances of the SPDNN, NPDNN and DAR predictors}
\label{tab1}
\end{table}

\section{Proofs of the main results}\label{prove}
\subsection{Proof of Theorem  \ref{thm1}}
For all $ h \in \mathcal{H}_{\sigma} (L, N, B, F, S),  ~ \varepsilon > 0, ~  \nu_0 \in (0, 1], ~ \lambda =  \dfrac{1}{n^{1 - \nu_0}} $  and for any $ n \in \N$,  in \cite{alquier2013prediction} and the Markov inequality we have 
\begin{align}\label{equa1}
\nonumber   P \Big\{ R (h) - \widehat{R}_n (h) > \varepsilon \Big\}  
 & = P \Big\{  \Big( \E [\ell (h (X_0), Y_0)] - \sum_{i = 1}^{n} \ell (h (X_i), Y_i) \Big) > n \varepsilon \Big\}
\\
\nonumber &  \leq    P \Big\{ \lambda \left |  \Big( \E [\ell (h (X_0), Y_0)] - \sum_{i = 1}^{n} \ell (h (X_i), Y_i) \Big) \right | >  n \lambda \varepsilon \Big\}
\\
\nonumber &  \leq   P \Big\{ e^{  \lambda |  ( \E [\ell (h (X_0), Y_0)] - \sum_{i = 1}^{n} \ell (h (X_i), Y_i) ) | } >  e^{n \lambda \varepsilon}  \Big\}
\\
& \leq \exp\Big ( - n^{\nu_0} \varepsilon +  \dfrac{  n^{2 \nu_0 - 1} (M + \theta_{\infty, n} (1))^2}{2 }\Big).
\end{align}
Set $ l = \mathcal{N} \Big ( \mathcal{H}_{\sigma} (L, N, B, F, S), \dfrac{\varepsilon}{4 G} \Big)$. We get from  (\ref{equa1}) and in \cite{diop2022statistical} for  $n \in \N$,
\begin{align}\label{equa2}
\nonumber  P \Big\{  \underset{h \in  \mathcal{H}_{\sigma} (L, N, B, F, S)}{\sup} \Big [R (h) - \widehat{R}_n (h) \Big] > \varepsilon \Big\} 
 & \leq \sum_{j =1}^l \exp\Big ( - n^{\nu_0} \varepsilon +  \dfrac{ n^{2 \nu_0 - 1} (M + \theta_{\infty, n} (1))^2}{2 }\Big)
\\
&  \leq  \mathcal{N} \Big ( \mathcal{H}_{\sigma} (L, N, B, F, S), \dfrac{\varepsilon}{4 G} \Big) \exp\Big ( - n^{\nu_0} \varepsilon +  \dfrac{ n^{2 \nu_0 - 1} (M + \theta_{\infty, n} (1))^2}{2 }\Big).
\end{align}
Which completes the proof of the Theorem \ref{thm1}.
\qed

\subsection{Proof of Theorem \ref{thm2}}
\indent (i) For    $ \varepsilon \in  (0,  2 M ] $, by using  (\ref{equa2}) and  the following inequality in \cite{ohn2019smooth}
\begin{equation}
\mathcal{N} \Big (\mathcal{H}_{\sigma} \left(L, N, B, F, S \right),  \dfrac{\varepsilon}{4 G} \Big)  \leq  \exp \Bigg (  2 L (S + 1) \log \left(\frac{4 G}{\varepsilon} C_{\sigma} L (N + 1)(B \lor 
 1) \right) \Bigg).
\end{equation}
We have 
\begin{align}\label{excessriskempirik}
\nonumber  &  P \Big\{  \underset{h \in  \mathcal{H}_{\sigma} (L, N, B, F, S)}{\sup} \Big [R (h) - \widehat{R}_n (h) \Big] > \varepsilon \Big\}
\\
\nonumber & \leq \exp \Bigg ( 2 L (S + 1) \log \left(\frac{4 G}{\varepsilon} C_{\sigma} L (N + 1)(B \lor 
 1) \right) \Bigg) \exp\Big ( - n^{\nu_0} \varepsilon +  \dfrac{ n^{2 \nu_0 - 1}  (M + \theta_{\infty, n} (1))^2}{2 }\Big)
\\
 &  \leq \exp \Bigg ( 2 L (S + 1) \log \left(\frac{4 G}{\varepsilon} C_{\sigma} L (N + 1)(B \lor 
 1)\right) - n^{\nu_0} \varepsilon +  \dfrac{ n^{2 \nu_0 -1} (M + \theta_{\infty, n} (1))^2}{2 }\Bigg) = \eta.
\end{align}
Let $0< \eta < 1$. We consider the following equation  with respect to $ \varepsilon$:
\begin{align}
\nonumber &  \exp \Bigg (  2 L (S + 1) \log \left(\frac{4 G}{\varepsilon} C_{\sigma} L (N + 1)(B \lor 
 1)\right) - n^{\nu_0} \varepsilon +  \dfrac{ n^{2 \nu_0 - 1} (M + \theta_{\infty, n} (1))^2}{2}\Bigg) = \eta,  
\end{align}
i.e.
\begin{align}
 2 L (S + 1) \log \left(\frac{4 G}{\varepsilon} C_{\sigma} L (N + 1)(B \lor 
 1)\right) - n^{\nu_0} \varepsilon +  \dfrac{ n^{2 \nu_0 - 1} (M + \theta_{\infty, n} (1))^2}{2 } = \log (\eta).
\end{align}
i.e.
\[  2 L (S + 1) \log \left( 4 G C_{\sigma} L (N + 1)(B \lor  1) \right)  -  2 L (S + 1)  \log (\varepsilon) - n^{\nu_0} \varepsilon +  \dfrac{ n^{2 \nu_0 - 1} (M + \theta_{\infty, n} (1))^2}{2 } -  \log (\eta) = 0.  \]
Set 
\[
C_{0, n} = n^{\nu_0},  ~   C_1 = 2 L (S + 1) \log \left( 4 G C_{\sigma} L (N + 1)(B \lor  1) \right).
\]
We have 
\begin{equation}
2 L (S + 1)  \log (\varepsilon) +  C_{0, n} \varepsilon -  \dfrac{ n^{2 \nu_0 - 1} (M + \theta_{\infty, n} (1))^2}{2 } +  \log (\eta)  - C_1= 0.
\end{equation}
Consider the function
\[
\varphi (\varepsilon) = 2 L (S + 1)  \log (\varepsilon) +  C_{0, n} \varepsilon -  \dfrac{ n^{2 \nu_0 - 1} (M + \theta_{\infty, n} (1))^2}{2 } +  \log (\eta)  - C_1 ~ \text{for} ~ \varepsilon \in (0,  2M].  \]
 We have,
$ \varphi (\varepsilon) \rightarrow - \infty ~ \text{as} ~ \varepsilon \rightarrow 0$.
Also,
\begin{align}\label{boundphi2M}
\nonumber \varphi (2 M)   > 0 & \Rightarrow  2 L (S + 1)  \log (2 M) +  2 C_{0, n} M -  \dfrac{ n^{2 \nu_0 - 1} (M + \theta_{\infty, n} (1))^2}{2 } +  \log (\eta)  - C_1   > 0
\\
\nonumber & \Rightarrow  2 C_{0, n} M -  \dfrac{ n^{2 \nu_0 - 1} (M + \theta_{\infty, n} (1))^2}{2 } >  C_1  - 2 L (S + 1)  \log (2 M) - \log (\eta) 
\\
\nonumber & \Rightarrow C_{0, n} -  \dfrac{ n^{2 \nu_0 - 1} (M + \theta_{\infty, n} (1))^2}{4 M } > \dfrac{1}{2 M} \Big ( C_1  - 2 L (S + 1)  \log (2 M) - \log (\eta) \Big)
\\
\nonumber & \Rightarrow n^{2 \nu_0}-  \dfrac{ n^{2 \nu_0 - 1} (M + \theta_{\infty, n} (1))^2}{4 M } > \dfrac{1}{2 M}\Big ( C_1  - 2 L (S + 1)  \log (2 M) - \log (\eta) \Big)
\\
 & \Rightarrow n^{2 \nu_0} \left( 1-  \dfrac{(M + \theta_{\infty, n} (1))^2}{4 M n} \right) > \dfrac{1}{2 M}\Big ( C_1  - 2 L (S + 1)  \log (2 M) - \log (\eta) \Big).
\end{align}
Since
$ \dfrac{(M + \theta_{\infty, n} (1))^2}{4 M n}  \longrightarrow 0$ as 
$n  \longrightarrow \infty $,
there exists $n_0 (\nu_0, \theta_{\infty, n}, M) $ such that,
\begin{equation}
n > n_0 (\nu_0, \theta_{\infty, n},  M) \Rightarrow \dfrac{(M + \theta_{\infty, n} (1))^2}{4 M n} < \dfrac{1}{2}.
\end{equation}
Thus, to get (\ref{boundphi2M}), with $n > n_0 (\nu_0, \theta_{\infty, n}, M) $ it suffices that,
\begin{align}
\dfrac{n^{2 \nu_0}}{2}  > \dfrac{1}{2 M}\Big ( C_1  - 2 L (S + 1)  \log (2 M) - \log (\eta) \Big).
\end{align}
Recall, $(x)_{+} = \max (x, 0) $ for all $x \in \R$.
That is,
\begin{equation}\label{bound_n}
n >  \Big( \dfrac{1}{M}\Big ( C_1  - 2 L (S + 1)  \log (2 M) - \log (\eta) \Big)_{+} \Big)^{1/(2 \nu_0)}.
\end{equation}
Therefore, the function $ \varepsilon \mapsto  \varphi (\varepsilon ) $ is strictly  increasing  and  under the condition (\ref{bound_n}), we get $ \varphi (\varepsilon) \rightarrow - \infty ~ \text{as} ~ \varepsilon \rightarrow 0$ and $ \varphi (2 M) > 0$. Hence, there exists a unique $ \varepsilon (n, \eta) \in (0, 2M)$, such that  $ \varphi \left( \varepsilon (n, \eta) \right) = 0$. 
Set $ \varepsilon_0 = \dfrac{2 M}{n^{\nu_0/2 }}, ~ \nu_0 < 1$, 
we have
\begin{align}\label{varepsipositive}
\nonumber \varphi (\varepsilon_0) > 0    & \Rightarrow  2 L (S + 1)  \log (2 M) -   L (S + 1) \nu_0 \log n +  \dfrac{2 MC_{0, n}}{n^{\nu_0  /2}} -  \dfrac{ n^{2 \nu_0 - 1} (M + \theta_{\infty, n} (1))^2}{2 } +  \log (\eta)  - C_1   > 0
\\
\nonumber & \Rightarrow  2M n^{\nu_0 /2}  -   L (S + 1) \nu_0 \log n -  \dfrac{ n^{2 \nu_0 - 1} (M + \theta_{\infty, n} (1))^2}{2 } >  C_1 -   2 L (S + 1)  \log (2 M) -  \log (\eta)
\\
 &  \Rightarrow 2M n^{2 \nu_0} \Big (1 -   L (S + 1) \nu_0 \dfrac{\log n}{2 M n^{2 \nu_0}} -   \dfrac{(M + \theta_{\infty, n} (1))^2}{4M n} \Big) >    C_1 -   2 L (S + 1)  \log (2 M) -  \log (\eta).
\end{align}
Since,
$
 L (S + 1) \nu_0 \dfrac{\log n}{2 M n^{\nu_0 /2}} +    \dfrac{(M + \theta_{\infty, n} (1))^2}{4M n }   \longrightarrow 0$ as
$n \rightarrow  \infty
$,
there exists $ n_0 \left( L, S, M, \nu_0, \theta_{\infty, n} \right) $ such that,
\begin{equation}\label{equan0}
n >  n_0 \left( L, S, M, \nu_0, \theta_{\infty, n} \right) \Rightarrow     L (S + 1) \nu_0 \dfrac{\log n}{2 M n^{\nu_0 /2}} +    \dfrac{(M + \theta_{\infty, n} (1))^2}{4M n } < \dfrac{1}{2}. 
\end{equation}
Thus, to get (\ref{varepsipositive}), with $ n >  n_0 \left( L, S, M, \nu_0, \theta_{\infty, n} \right) $  it suffice that, 
\[    M n^{2\nu_0} >  C_1 -   2 L (S + 1)  \log (2 M) -  \log (\eta).  \]
i.e.
\begin{equation}\label{bound_n2}
 n >   \Big (  \dfrac{1}{ M} \left ( C_1 -   2 L (S + 1)  \log (2 M) -  \log (\eta) \right)_{+} \Big)^{1/2\nu_0}. 
 \end{equation}
Thus, under the condition (\ref{bound_n2}), the unique solution $ \varepsilon_1 \left (n, \eta, \nu_0 \right) $ of $ \varphi \left(\varepsilon (n, \eta) \right) = 0$, satisfies $ \varepsilon_1 \left (n, \eta, \nu_0 \right)  < \dfrac{2M}{n^{\nu_0/2 }} < 2 M$ for $ \nu_0 < 1 ) $.
Therefore, from (\ref{excessriskempirik}),  it holds that, with probability at least $ 1 - \eta$,
\[    \underset{h \in  \mathcal{H}_{\sigma} (L, N, B, F, S)}{\sup} \Big [R (h) - \widehat{R}_n (h) \Big] \leq  \varepsilon_1 \left (n, \eta, \nu_0 \right);   \]
which implies,
\begin{equation}\label{excessemp}
R (\widehat{h}_{n,NP}) - \widehat{R}_n (\widehat{h}_{n,NP}) \leq  \varepsilon_1 \left (n, \eta, \nu_0 \right).
\end{equation}
The first attempt of the  Theorem \ref{thm2} hold.

\medskip

\textbf{(ii)}  From (\ref{equa1}), we have,
\begin{equation}\label{equ_bound_target_function}
P \Big\{\widehat{R}_n (h_{\mathcal{H}_{\sigma}}) -  R (h_{\mathcal{H}_{\sigma}} ) > \varepsilon \Big\} \leq  \exp\Big ( - n^{\nu_0} \varepsilon +  \dfrac{  n^{2 \nu_0 - 1} (M + \theta_{\infty, n} (1))^2}{2 }\Big) = \eta.
\end{equation}
Let $0 < \eta < 1 $. Consider the equation with respect to $ \varepsilon$,
 \[ \exp\Big ( - n^{\nu_0} \varepsilon +  \dfrac{  n^{2 \nu_0 - 1} (M + \theta_{\infty, n} (1))^2}{2 } \Big) = \eta.  \]
i.e.
\begin{align}
- n^{\nu_0} \varepsilon +  \dfrac{  n^{2 \nu_0 - 1} (M + \theta_{\infty, n} (1))^2}{2 } - \log (\eta) = 0.
\end{align}
A solution of this equation is,
\[ \varepsilon'_1 (n, \eta, \nu_0) = \dfrac{(M + \theta_{\infty, n} (1))^2}{2 n^{1 - \nu_0}} - \dfrac{\log (\eta)}{n^{\nu_0}}.       \]
Thus, from (\ref{equ_bound_target_function}), we have with a probability at least $ 1 - \eta$,
\begin{equation}\label{excessrisq}
\widehat{R}_n (h_{\mathcal{H}_{\sigma}}) -  R (h_{\mathcal{H}_{\sigma}} ) \leq  \varepsilon'_1 (n, \eta, \nu_0).
\end{equation}
Recall that $ \widehat{h}_n = \underset{h \in \mathcal{H}_{\sigma} \left(L, N, B, F, S \right)}{\argmin} \widehat{R}_n (h) $. Since $ h_{\mathcal{H}_{\sigma}} \in  \mathcal{H}_{\sigma} \left(L, N, B, F, S \right) $,  we have 
\begin{equation}
 \widehat{R}_n (\widehat{h}_{n,NP}) \leq \widehat{R}_n (h_{\mathcal{H}_{\sigma}}).
\end{equation}
We deduce, 
\begin{equation}\label{excessrisq}
\widehat{R}_n (h_{\mathcal{H}_{\sigma}}) -  R (h_{\mathcal{H}_{\sigma}} ) \leq  \varepsilon'_1 (n, \eta, \nu_0).
\end{equation}
Combining (\ref{excessemp}) and (\ref{excessrisq}), with probability at least $ 1 - 2 \eta$, we get
\begin{align}
R (\widehat{h}_{n,NP}) - R (h_{\mathcal{H}_{\sigma}} ) \leq  \varepsilon_1 (n, \eta, \nu_0) + \varepsilon'_1 (n, \eta, \nu_0).
\end{align}
Which completes the proof of the Theorem \ref{thm2}.
\qed

\subsection{Proof of Theorem \ref{thm4}}
Let us consider the following decomposition:
\[ R(\widehat{h}_n) - R(h^{*})   \coloneqq B_{1, n} +  B_{2, n}. \]
Where,
\[
\begin{array}{llll}
  B_{1, n}   & =   [R(\widehat{h}_n) - R(h^{*})]  - 2 [ \widehat{R}_n (\widehat{h}_n)  - \widehat{R}_n (h^{*}) ] - 2 J_{\lambda_n, \tau_n} (\widehat{h}_n); 
  \\
 B_{2, n}   &  =   2[\widehat{R}_n (\widehat{h}_n) - \widehat{R}_n (h^{*})] + 2 J_{\lambda_n, \tau_n} (\widehat{h}_n).
\end{array}
\]
For bound $B_{1, n}$  using similarly way   in  \cite{kengne2023sparse}.
Let 

\[\Delta (h) (Z_0) \coloneqq \ell (h (X_0), Y_0) - \ell (h^{*} (X_0), Y_0) ~ \text{ with} ~ Z_0  \coloneqq (X_0, Y_0).     \]
Let us defined 
\[ \mathcal{H}_{n, j, \rho} \coloneqq  \left\{ h \in \mathcal{H}_{\sigma}(L_n, N_n, B_n, F): 2^{j - 1} \ind_ {\{j \ne 0 \} } \rho  \leq J_{\lambda_n, \tau_n} (h) \leq  2^{j} \rho \right\}.    \]
 Thus for $\rho > 0$,  we have
\begin{align}
\nonumber    P(B_{1, n} > \rho)   &  = P \Big ( \E [\ell ( \widehat{h}_n (X_0), Y_0)] - \E[ \ell (h^{*} (X_0), Y_0)] - \dfrac{2}{n} \sum_{i= 1}^n \left[ \ell (\widehat{h}_n (X_i), Y_i) -  \ell ((h^{*} (X_i), Y_i)  \right] - 2 J_{\lambda_n, \tau_n} (\widehat{h}_n) >  \rho \Big)
\\
\nonumber & \leq P\Big( \E [\ell ( h (X_0), Y_0)] - \E[ \ell (h^{*} (X_0), Y_0)] - \dfrac{1}{n} \sum_{i= 1}^n \left[ \ell ( h (X_i), Y_i) -  \ell (h^{*} (X_i), Y_i)  \right] > \dfrac{1}{2}  \Big( \E [\ell ( h (X_0), Y_0)] 
\\
\nonumber & \hspace{10cm} - \E[ \ell (h^{*} (X_0), Y_0)]   + 2 J_{\lambda_n, \tau_n} (h) + \rho  \Big) \Big)  
\\
\nonumber &  \leq P \Big( \exists h \in \mathcal{H}_{\sigma} (L_n, N_n, B_n, F): \E [\Delta (h) (Z_0)] - \dfrac{1}{n} \sum_{i= 1}^{n} \E [\Delta (h) (Z_i)] >  \dfrac{1}{2} \Big( \rho + 2 J_{\lambda_n, \tau_n} (h) + \E [\Delta (h) (Z_0)] \Big)   \Big)
\\
 &  \leq  \sum_{j= 0}^{\infty} P \Bigg( \underset{h \in \mathcal{H}_{n, j, \rho}}{\sup} \dfrac{\E [\Delta (h) (Z_0)] - \dfrac{1}{n} \sum_{i= 1}^{n} \E [\Delta (h) (Z_i)]}{\rho + 2 J_{\lambda_n, \tau_n} (h) + \E [\Delta (h) (Z_0)] } >  \dfrac{1}{2}    \Bigg).
\end{align}

Since,
\[\E [\Delta (h) (Z_0)] \ge  0, ~ \rho  >0, ~ \text{and for} ~ j \ne 0, 2 J_{\lambda_n, \tau_n} \geq 2^j \rho.    \]
Thus we have
\begin{align}
\nonumber &   P(B_{1, n} > \rho)     \leq \sum_{j=1}^{\infty}  P \Big( \underset{h \in \mathcal{H}_{n, j, \rho}}{\sup} \E [\Delta (h) (Z_0)] - \dfrac{1}{n} \sum_{i= 1}^{n} \E [\Delta (h) (Z_i)] >  \dfrac{2^j \rho}{2}    \Big). 
\end{align}
Let us defined by
\begin{equation}
 \mathcal{G}_{n, j, \rho} \coloneqq \left\{ \Delta (h): \R^d \times \mathcal{Y} \rightarrow  \R: h \in \mathcal{H}_{n, j, \rho}  \right\}.
\end{equation}
Since, $ h \in \mathcal{H}_{\sigma}(L_n, N_n, B_n, F), ~ h^{*} \in \mathcal{C}^{s, \mathcal{K}}$ are Lipschitzian with Lipschitz coefficient denote  respectively by 
 $\mathcal{K}_{\mathcal{H}_{\sigma}(L_n, N_n, B_n, F)}, \mathcal{K}_{\mathcal{C}^{s, \mathcal{K}}} > 0 $.
Let $ g(x, y) =  \Delta (h)(x, y)$, on can see that   for  $ (x, y), ~ (x', y') \in \R^d \times \R$, 
\begin{align}
\nonumber | g(x,  y) - g(x', y')| &  = \left|\Delta (h)(x, y) - \Delta (h)(x', y') \right|   = \left| [\ell (h(x), y) - \ell (h(x'), y')] - [\ell (h^{*} (x), y) - \ell (h^{*} (x'), y') ]  \right|
\\
\nonumber & \leq \left| \ell (h(x), y) - \ell (h(x'), y')  \right| + \left| \ell (h^{*} (x), y) - \ell (h^{*} (x'), y')   \right|
\\
\nonumber & \leq \mathcal{K}_{\ell} \Big( \left|  (h(x), y) -  (h(x'), y')  \right| + \left|  (h^{*} (x), y) -  (h^{*} (x'), y')   \right| \Big)
\\
\nonumber & \leq \mathcal{K}_{\ell} \Big( \left|  h(x) -  h(x') \right| + \left|  h^{*} (x)-  h^{*} (x')   \right| + 2 |y  -  y'| \Big)
\\
\nonumber & \leq \mathcal{K}_{\ell} \Big( \mathcal{K}_{\mathcal{H}_{\sigma}(L_n, N_n, B_n, F)}\left|  x -  x' \right| +  \mathcal{K}_{\mathcal{C}^{s, \delta}} \left| x-  x' \right| + 2 |y  -  y'| \Big)
\\
& \leq  2 \mathcal{K}_{\ell}  \max  (\mathcal{K}_{\mathcal{H}_{\sigma} (L_n, N_n, B_n, F)} + \mathcal{K}_{\mathcal{C}^{s, \mathcal{K}}}, 1) \Big(   \|  x -  x'\|  +  |y  -  y'| \Big).
\end{align}
Thus,  the function $g$ is $  2 \mathcal{K}_{\ell}  \max  (\mathcal{K}_{\mathcal{H}_{\sigma} (L_n, N_n, B_n, F)} + \mathcal{K}_{\mathcal{C}^{s, \mathcal{K}}}, 1)$-Lipschitz. Hence  the process $\{ g(Z_t) \coloneqq \ell (h(X_t), Y_t), t \in \Z 
 \}$ is  $\psi$-weakly dependent. Thus, we have
\begin{align}
 \nonumber    P(B_{1, n}   > \rho)   &  \leq \sum_{j=1}^{\infty}  P \Bigg( \underset{h \in \mathcal{H}_{n, j, \rho}}{\sup} \E [g (Z_0)] - \dfrac{1}{n} \sum_{i= 1}^{n} \E [g (Z_i)] >  \dfrac{2^j \rho}{2} \Bigg)   
 \\
 & \leq  \sum_{j=1}^{\infty}  P \Bigg( \underset{g \in \mathcal{G}_{n, j, \rho}}{\sup} \E [g (Z_0)] - \dfrac{1}{n} \sum_{i= 1}^{n} \E [g(Z_i)] >  \dfrac{2^j \rho}{2}    \Bigg). 
\end{align}

One can see that 
\begin{align}\label{equ_b}
\nonumber   P\Big\{\E[g (Z_0)] - \dfrac{1}{n} \sum_{i=1}^n g (Z_i) > \varepsilon \Big\}  & =  P\Big\{  \Big (\E[g (X_0, Y_0)] -  \sum_{i=1}^n g (X_i, Y_i) \Big) \ge  n \varepsilon  \Big\} 
 \\
   & \leq P\Big\{ \Big| \Big(\E [g (X_0, Y_0)] -  \sum_{i=1}^n g (X_i, Y_i)  \Big) \Big| \ge  
 n \varepsilon \Big\}.  
 \end{align}

 Let $l = \mathcal{N}( \varepsilon, \mathcal{G}_ {n, j, \rho}, \| \cdot \|_\infty) $.  Recall the conditions
 \[ A_n \ge \E \Big[ \Big( \sum_{i = 1}^{n} \Big( \ell (h(X_i), Y_i) -  \E [\ell(h(X_0), Y_0)] \Big) \Big)^{2} \Big],  \quad B_n = 2 ML_2 \max \Bigg( \dfrac{2^{4 + \mu} n M^2 L_1}{A_n}, 1\Bigg).   \]
 One can see in \cite{doukhan2007probability} and \cite{kengne2023deep}
\begin{align}\label{equ_br}
 P \Big\{ \underset{ g \in \mathcal{G}_{n, j, \rho}}{\sup} \Big[ \E[g(Z_0)] - \dfrac{1}{n} \sum_{i= 1}^n g(Z_i) \Big] > \varepsilon \Big\}  & \leq \mathcal{N} \left(\varepsilon, \mathcal{G}_{n, j, \rho}, \| \cdot \|_{\infty} \right) \exp \Bigg ( - \dfrac{n^2 \varepsilon^2/8}{ A_n + B_n^{1/ (\mu + 2)} (n \varepsilon/2)^{(2 \mu + 3)/(\mu + 2)} } \Bigg).
\end{align}
 In \cite{ohn2022nonconvex}, we have  
\begin{equation}\label{P_inqu2}
\mathcal{N}(\varepsilon, \mathcal{G}_{n, j, \rho}, \| \cdot \|_\infty) \leq \mathcal{N} \left(\frac{\varepsilon}{\mathcal{K_{\ell}}}, \mathcal{H}_{n, j, \rho}, \| \cdot \|_\infty \right).
\end{equation} 
One can easily see that,
\begin{equation}\label{inclusion}
 \mathcal{H}_{n, j, \rho}  \subset \left\{  h \in \mathcal{H}_{\sigma}(L_{n}, N_{n}, B_{n}, F, \dfrac{2^j \rho}{\lambda_n}): \| \theta(h) \|_{ \text{clip}, \tau_n} \leq \frac{2^j \rho}{\lambda_n}  \right\}. 
 \end{equation}
Let $ \Delta (h_1), ~  \Delta (h_2) \in \mathcal{G}_{n, j, \rho}, ~ \text{and} ~ (x, y) \in \R^d \times \R$, we have
\begin{align}
\nonumber \| \Delta (h_1) (x, y) -  \Delta (h_2) (x, y) \|_{\infty} & = \left| \ell (h_1 (x), y) - \ell (h_2 (x), y) \right| 
\\
&  \leq  \mathcal{K}_{\ell} \left| h_1 (x) - h_2 (x) \right|.
\end{align}
One can see in \cite{ohn2022nonconvex} the following inequality
\begin{align}\label{cover_number}
\mathcal{N} \left( \varepsilon, \mathcal{G}_{n, j, \rho}, \| \cdot \|_\infty \right)  \nonumber & \leq \mathcal{N} \Big( \dfrac{\varepsilon}{\mathcal{K}_{\ell}}, \mathcal{H}_{n, j, \rho}, \| \cdot \|_\infty \Big) 
\leq \mathcal{N} \Big( \frac{\varepsilon}{\mathcal{K}_{\ell}}, \mathcal{H}_{\sigma}(L_{n},N_{n}, B_{n}, F, \frac{2^j \rho}{\lambda_n}),  \| \cdot \|_\infty \Big)
\\
& \leq  \exp\left( 2 \frac{2^j \rho}{\lambda_n}(L_n + 1) \log \left(\frac{(L_n + 1)(N_n + 1)B_n}{\dfrac{\varepsilon}{\mathcal{K}_{\ell}} - \tau_n (L_n + 1)((N_n + 1) B_n)^{L_n +1}} \right) \right).
\end{align} 
Thus, 
\begin{align}\label{equ_B1}
\nonumber  & P\Big\{\underset{ g \in \mathcal{G}_{n, j, \rho}}{\sup} \Big[ \E[g(Z_0)] - \dfrac{1}{n} \sum_{i= 1}^n g(Z_i) \Big] > \varepsilon \Big\} 
\\
\nonumber & \leq  \exp\left(2 \frac{2^j \rho}{\lambda_n}(L_n + 1) \log \left(\frac{(L_n + 1)(N_n + 1)B_n}{\dfrac{\varepsilon}{\mathcal{K}_{\ell}} - \tau_n (L_n + 1)((N_n + 1) B_n)^{L_n +1}} \right) \right)   \exp \Bigg ( - \dfrac{n^2 \varepsilon^2/8}{ A_n + B_n^{1/(\mu + 2) } (n \varepsilon/2)^{(2 \mu + 3)/(\mu + 2)} } \Bigg)
\\
& \leq   \exp\left(2 \frac{2^j \rho}{\lambda_n}(L_n + 1) \log \left(\frac{(L_n + 1)(N_n + 1)B_n}{\dfrac{\varepsilon}{\mathcal{K}_{\ell}} - \tau_n (L_n + 1)((N_n + 1) B_n)^{L_n +1}} \right)  - \dfrac{n^2 \varepsilon^2/8}{ A_n + B_n^{1/(\mu + 2) } (n \varepsilon/2)^{(2 \mu + 3)/(\mu + 2)} } \right). 
\end{align}
Let $ \varepsilon = \dfrac{2^j \rho}{2}$,  we have
\begin{align}\label{equ_B2}
\nonumber  & P\Big\{\underset{ g \in \mathcal{G}_{n, j, \rho}}{\sup} \Big[ \E[g(Z_0)] - \dfrac{1}{n} \sum_{i= 1}^n g(Z_i) \Big] > \dfrac{2^j \rho}{2} \Big\} 
\\
 & \leq  \exp\left(2 \frac{2^j \rho}{\lambda_n}(L_n + 1) \log \left(\frac{(L_n + 1)(N_n + 1)B_n}{\dfrac{2^j  \rho}{2\mathcal{K}_{\ell}} - \tau_n (L_n + 1)((N_n + 1) B_n)^{L_n +1}} \right)   - \dfrac{n^2 (2^j \rho)^2/32}{ A_n + B_n^{1/(\mu + 2) } (n 2^j \rho/4)^{(2 \mu + 3)/(\mu + 2)} }\right).
\end{align}
With a choice of $ A_n, B_n, C_{1, n}, C_{2, n}$ as in \cite{kengne2023deep} we have
\begin{align}\label{equ_B2}
\nonumber  & P\Big\{\underset{ g \in \mathcal{G}_{n, j, \rho}}{\sup} \Big[ \E[g(Z_0)] - \dfrac{1}{n} \sum_{i= 1}^n g(Z_i) \Big] > \dfrac{2^j \rho}{2} \Big\} 
\\
 & \leq  \exp\left(2 \frac{2^j \rho}{\lambda_n}(L_n + 1) \log \left(\frac{(L_n + 1)(N_n + 1)B_n}{\dfrac{2^j  \rho}{2\mathcal{K}_{\ell}} - \tau_n (L_n + 1)((N_n + 1) B_n)^{L_n +1}} \right)   - \dfrac{n^2 (2^j \rho)^2/16}{ C_{1, n} n + 2 C_{2, n}^{1/(\mu + 2 )} (n 2^j \rho/4)^{(2 \mu + 3)/(\mu + 2)} }\right).
\end{align}
Let,
\begin{align}\label{equarhon}
\nonumber 2 C_{2, n}^{1/(\mu + 2) } (n 2^j \rho/4)^{(2 \mu + 3)/(\mu + 2)} & > C_{1, n} n, 
\\
 \Longrightarrow \rho & >  \dfrac{\left( C_{1, n}/(2 C_{2, n}^{1/(\mu + 2)})\right)^{(\mu + 2)/(2 \mu + 3)}}{n^{(\mu + 1) / (2\mu + 3)}} \coloneqq \rho_n.
\end{align}
One can see from (\ref{equarhon}) that 
\[- \dfrac{n^2 (2^j \rho)^2/16}{ C_{1, n} n + 2 C_{2, n}^{1/(\mu + 2) } (n 2^j \rho/4)^{(2 \mu + 3)/(\mu + 2)} } \leq - \dfrac{n^2 (2^j \rho)^2/16}{ 4C_{2, n}^{1/ (\mu + 2) } (n 2^j \rho/4)^{(2 \mu + 3)/ (\mu + 2)}}.   \]
\textbf{Step 1}: $ \rho > 1 > \rho_n $. Recall the condition
\[ \tau_n  \leq \dfrac{1}{ 4 \mathcal{K}_{\ell} (L_n + 1)((N_n + 1) B_n)^{L_n +1}}.     \]
We have
\begin{align}\label{equ_B2}
\nonumber  & P\Big\{\underset{ g \in \mathcal{G}_{n, j, \rho}}{\sup} \Big[ \E[g(Z_0)] - \dfrac{1}{n} \sum_{i= 1}^n g(Z_i) \Big] > \dfrac{2^j \rho}{2} \Big\} 
\\
 & \leq  \exp\left(2 \frac{2^j \rho}{\lambda_n}(L_n + 1) \log \left(\frac{(L_n + 1)(N_n + 1)B_n}{\dfrac{2^j  \rho}{2\mathcal{K}_{\ell}} - \tau_n (L_n + 1)((N_n + 1) B_n)^{L_n +1}} \right)   - \dfrac{n^2 (2^j \rho)^2/16}{ 4C_{2, n}^{1/ (\mu + 2)} (n 2^j \rho/4)^{(2 \mu + 3)/(\mu + 2)}}  \right).
\end{align}
Since  $ 2^j \rho > 1$
\begin{align}\label{equa_taun}
\nonumber \dfrac{2^j \rho}{4 \mathcal{K}_{\ell}} > \dfrac{1}{4 \mathcal{K}_{\ell}} & \Longrightarrow  \dfrac{2^j \rho}{2 \mathcal{K}_{\ell}} - \dfrac{1}{4 \mathcal{K}_{\ell}} > \dfrac{2^j \rho}{2 \mathcal{K}_{\ell}} - \dfrac{2^j \rho}{4 \mathcal{K}_{\ell}} = \dfrac{2^j \rho}{4 \mathcal{K}_{\ell}}
\\
 & \Longrightarrow \dfrac{1}{\dfrac{2^j \rho}{2 \mathcal{K}_{\ell}} - \dfrac{1}{4 \mathcal{K}_{\ell}}} < \dfrac{4 \mathcal{K}_{\ell}}{ 2^j \rho}
\Longrightarrow \dfrac{(L_n + 1)(N_n + 1)B_n}{\dfrac{2^j \rho}{2 \mathcal{K}_{\ell}} - \dfrac{1}{4 \mathcal{K}_{\ell}}} < \dfrac{4 \mathcal{K}_{\ell} (L_n + 1)(N_n + 1)B_n}{ 2^j \rho}.
\end{align}
Thus from (\ref{equa_taun}),  and by  applying the inequality $ \log x < m (x^{1/m} -1), \quad \text{for all} ~ x > 0, ~ m >0 $, we have
\begin{align}\label{equ_B2}
\nonumber & P\Big\{\underset{ g \in \mathcal{G}_{n, j, \rho}}{\sup} \Big[ \E[g(Z_0)] - \dfrac{1}{n} \sum_{i= 1}^n g(Z_i) \Big] > \dfrac{2^j \rho}{2} \Big\} 
\\
\nonumber & \leq  \exp\left(2 \frac{2^j \rho}{\lambda_n}(L_n + 1) m \left(  \left(\dfrac{4 \mathcal{K}_{\ell} (L_n + 1)(N_n + 1)B_n}{ 2^j \rho} \right)^{1/m} - 1 \right) -  \dfrac{n^2 (2^j \rho)^2/16}{ 4C_{2, n}^{1/ (\mu + 2)} (n 2^j \rho/4)^{(2 \mu + 3)/(\mu + 2)}} \right)
\\
& \leq  \exp\left( 2\dfrac{(L_n + 1) m}{\lambda_n} (2^j \rho)^{1 - 1/m}  \left( 4 \mathcal{K}_{\ell} (L_n + 1) (N_n + 1) B_n \right)^{1/m} - 2 \frac{2^j \rho}{\lambda_n} (L_n + 1)m   -  \dfrac{n^{1/(\mu + 2)} (2^j \rho)^{1/ (\mu + 2)}/16}{ 4C_{2, n}^{1/\mu + 2 } (1/4)^{(2 \mu + 3) /(\mu + 2)}} \right).
\end{align}
Let,
\[1 - \dfrac{1}{m} =  \dfrac{1}{\mu + 2} \Longleftrightarrow m = \dfrac{\mu + 2}{\mu + 1}.  \]
Thus,
\begin{align}\label{equ_B2}
\nonumber & P\Big\{\underset{ g \in \mathcal{G}_{n, j, \rho}}{\sup} \Big[ \E[g(Z_0)] - \dfrac{1}{n} \sum_{i= 1}^n g(Z_i) \Big] > \dfrac{2^j \rho}{2} \Big\} 
\\
\nonumber & \leq  \exp\Bigg( 2\dfrac{(L_n + 1) (\mu + 2)}{\lambda_n(\mu + 1)} (2^j \rho)^{1/(\mu + 2)}  \left( 4 \mathcal{K}_{\ell} (L_n + 1) (N_n + 1) B_n \right)^{\mu + 1/(\mu + 2)} - 2 \frac{2^j \rho (L_n + 1) (\mu + 2)}{\lambda_n(\mu + 1)} \\
\nonumber & \hspace{9cm} -  \dfrac{n^{1/(\mu + 2)} (2^j \rho)^{1/ (\mu + 2)}/16}{ 4C_{2, n}^{1/\mu + 2 } (1/4)^{(2 \mu + 3) /(\mu + 2)}} \Bigg)
\\
\nonumber  & \leq  \exp\Bigg( 2\dfrac{(L_n + 1) (\mu + 2)}{\lambda_n(\mu + 1)} (2^j \rho)^{1/(\mu + 2)}  \left( 4 \mathcal{K}_{\ell} (L_n + 1) (N_n + 1) B_n \right)^{\mu + 1/ (\mu + 2)} -   \dfrac{n^{1/(\mu + 2)} (2^j \rho)^{1/ (\mu + 2)}/16}{ 4C_{2, n}^{1/\mu + 2 } (1/4)^{(2 \mu + 3) /(\mu + 2)}}  \Bigg)
\\
 & \leq  \exp\Bigg(  (2^j \rho)^{1/(\mu + 2)} \left( 2\dfrac{(L_n + 1) (\mu + 2)}{\lambda_n(\mu + 1)}   \left( 4 \mathcal{K}_{\ell} (L_n + 1) (N_n + 1) B_n \right)^{\mu + 1/ (\mu + 2)} -   \dfrac{n^{1/(\mu + 2)} /16}{ 4C_{2, n}^{1/\mu + 2 } (1/4)^{(2 \mu + 3) /(\mu + 2)}} \right)  \Bigg).
\end{align}
Recall the following  assumptions 
\begin{equation}\label{condi_of_seq}
 L_n \lesssim \log n, ~ N_n \lesssim n^{\nu_1}, B_n \lesssim n^{\nu_2}, ~ \lambda_n \asymp (\log n)^{\nu_3}/n^{\nu_4 }, ~ \nu_1 + \nu_2 + \nu_4 < \dfrac{1}{\mu + 2} \quad \text{for some} ~ \nu_1, ~  \nu_2, ~ \nu_3, ~  \nu_4 > 0.   
\end{equation}
Under the conditions on (\ref{condi_of_seq}) above,   there exists  $ N_1 \in \N$ such that if $ n > N_1$ we have
\begin{equation}
  2\dfrac{(L_n + 1) (\mu + 2)}{\lambda_n(\mu + 1)}   \left( 4 \mathcal{K}_{\ell} (L_n + 1) (N_n + 1) B_n \right)^{\mu + 1/ (\mu + 2)} <    \dfrac{n^{1/(\mu + 2)} /16}{ 8 C_{2, n}^{1/\mu + 2 } (1/4)^{(2 \mu + 3) /(\mu + 2)}}. 
\end{equation}
Thus  we have
\begin{equation}\label{equ_B5}
P\Big\{\underset{ g \in \mathcal{G}_{n, j, \rho}}{\sup} \Big[ \E[g(Z_0)] - \dfrac{1}{n} \sum_{i= 1}^n g(Z_i) \Big] > \dfrac{2^j \rho}{2} \Big\} 
\leq \exp\left( - (2^j \rho)^{1/(\mu + 2)} \dfrac{n^{1/(\mu + 2)} /16}{ 8 C_{2, n}^{1/\mu + 2 } (1/4)^{(2 \mu + 3) /(\mu + 2)}}\right).
\end{equation}

\begin{equation}\label{equ_B5}
\sum_{j = 1}^{\infty} P\Big\{\underset{ g \in \mathcal{G}_{n, j, \rho}}{\sup} \Big[ \E[g(Z_0)] - \dfrac{1}{n} \sum_{i= 1}^n g(Z_i) \Big] > \dfrac{2^j \rho}{2} \Big\} 
\leq  \sum_{j = 1}^{\infty} \exp\left( - (2^{1/(\mu + 2)})^j  \dfrac{n^{1/(\mu + 2)} /16}{ 8 C_{2, n}^{1/\mu + 2 }(1/4)^{(2 \mu + 3) /(\mu + 2) }} \rho^{1/(\mu + 2)} \right).
\end{equation}
Let,
\[ \beta_n \coloneqq  \dfrac{n^{1/(\mu + 2)} /16}{ 8 C_{2, n}^{ (1/\mu + 2 )} (1/4)^{(2 \mu + 3) /(\mu + 2)}} \rho^{1/(\mu + 2)},   \]
We have 
\begin{equation}\label{boundbetan}
\sum_{j =0}^{\infty} \exp \left(- \beta_n(2^{1/(\mu + 2)})^j \right)  \leq \exp \left( - \beta_n \right) \sum_{j = 0}^{\infty} \left( (2^{1/(\mu + 2)})^{-\frac{\beta_n}{\log 2}} \right)^j \leq \exp \left( - \beta_n \right) \Bigg ( \dfrac{1}{1 - (2^{1/(\mu + 2)})^{-\frac{\beta_n}{\log 2}}} \Bigg) \leq  2 \exp \left( - \beta_n \right).
\end{equation}
Hence,
\begin{equation}
P( B_{1, n} > \rho ) \leq 2 \exp\left( -  \dfrac{n^{1/(\mu + 2)} /16}{ 8 C_{2, n}^{1/\mu + 2 } (1/4)^{(2 \mu + 3) /(\mu + 2)}} \rho^{1/(\mu + 2)} \right).
\end{equation}
Thus, for  $ \rho > 1 $ we have
\begin{align}
\nonumber \displaystyle\int_{1}^{\infty} P( B_{1, n} > \rho ) d \rho  & \leq 
 2 \displaystyle\int_{1}^{\infty} \exp\left( -  \dfrac{n^{1/(\mu + 2)} /16}{ 8 C_{2, n}^{1/\mu + 2 } (1/4)^{(2 \mu + 3) /(\mu + 2)}} \rho^{1/(\mu + 2)} \right) d \rho.
\end{align}
Let
\[ x = \dfrac{n^{1/(\mu + 2)} /16}{ 8 C_{2, n}^{1/\mu + 2 } (1/4)^{(2 \mu + 3) /(\mu + 2)}} \rho^{1/(\mu + 2)},  ~ g(x) \coloneqq x - 2(\mu + 2) \log(x). \]
Let $ x_0 = (2(\mu + 2))^3 $.
One can  easily see  that  $ g(x_0) > 0$ and  for obtaining
\[ x > x_0  ~ \text{that is} \quad \dfrac{n^{1/(\mu + 2)} /16}{ 8 C_{2, n}^{1/\mu + 2 } (1/4)^{(2 \mu + 3) /(\mu + 2)}} \rho^{1/(\mu + 2)} >  (2(\mu + 2))^3, \quad \forall \rho > 1,  \]
it suffice that,
\[
n > \left( 128 C_{2, n}^{1/\mu + 2 } (1/4)^{(2 \mu + 3) /(\mu + 2)} (2(\mu + 2))^3 \right)^{\mu + 2}.
\]
One can remark that for any 
\begin{equation}\label{boundPb1}
x > x_0, \quad \text{we have} \quad  \exp(- x) <  \dfrac{1}{x^{2(\mu + 2)}}.
\end{equation}
One can see from (\ref{boundPb1})
\begin{align}
\nonumber \displaystyle\int_{1}^{\infty} P( B_{1, n} > \rho ) d \rho  & \leq 
 2 \displaystyle\int_{1}^{\infty} \exp\left( -  \dfrac{n^{1/(\mu + 2)} /16}{ 8 C_{2, n}^{1/\mu + 2 } (1/4)^{(2 \mu + 3) /(\mu + 2)}} \rho^{1/(\mu + 2)} \right) d \rho 
 \\
 \nonumber & \leq   2 \displaystyle\int_{1}^{\infty}  \dfrac{1}{\left(   \dfrac{n^{1/(\mu + 2)} /16}{ 8 C_{2, n}^{1/\mu + 2 } (1/4)^{(2 \mu + 3) /(\mu + 2)}} \rho^{1/(\mu + 2)} \right)^{2(\mu + 2)}}  d \rho 
 \\
 \nonumber & \leq  2\dfrac{(128)^{(2(\mu + 2))} C_{2, n}^2 (1/4)^{(2(2 \mu + 3))}}{n^2} \displaystyle\int_{1}^{\infty}   \dfrac{1}{\rho^2}d \rho 
 \\
 & \leq 2 \dfrac{(128)^{(2(\mu + 2))} (1/4)^{(2(2 \mu + 3))} C_{2, n}^2 }{n^2}.
\end{align}
\textbf{Step 2}: $ \rho_n < \rho < 1$.
Recall the condition
\[ \tau_n  \leq  \dfrac{\rho_n}{4\mathcal{K}_{\ell}(L_n + 1) ((N_n + 1)B_n)^{L_n + 1}}.   \]
\noindent Since $\rho  >  \rho_n$  we have  also
\[ \tau_n  \leq \dfrac{\rho}{4\mathcal{K}_{\ell}(L_n + 1) ((N_n + 1)B_n)^{L_n + 1}}.   \]
Thus
\begin{align}\label{equ_B2}
\nonumber  & P\Big\{\underset{ g \in \mathcal{G}_{n, j, \rho}}{\sup} \Big[ \E[g(Z_0)] - \dfrac{1}{n} \sum_{i= 1}^n g(Z_i) \Big] > \dfrac{2^j \rho}{2} \Big\} 
\\
\nonumber  & \leq  \exp\left(2 \frac{2^j \rho}{\lambda_n}(L_n + 1) \log \left(\frac{(L_n + 1)(N_n + 1)B_n}{\dfrac{2^j  \rho}{2\mathcal{K}_{\ell}} - \tau_n (L_n + 1)((N_n + 1) B_n)^{L_n +1}} \right)   - \dfrac{n^2 (2^j \rho)^2/16}{ 4C_{2, n}^{1/\mu + 2 } (n 2^j \rho/4)^{(2 \mu + 3)/(\mu + 2)}}  \right)
\\
& \leq  \exp\left(2 \frac{2^j }{\lambda_n}(L_n + 1) \log \left(\frac{(L_n + 1)(N_n + 1)B_n}{ \dfrac{ \rho}{ 2\mathcal{K}_{\ell}} \left(2^j - \dfrac{1}{ 2} \right) } \right)   - \dfrac{n^{1/(\mu + 2)} (2^j \rho)^{1/(\mu + 2)}/16}{ 4C_{2, n}^{1/\mu + 2 } (1/4)^{(2 \mu + 3)/(\mu + 2)}}  \right).
\end{align}
Thus we have
\begin{align}\label{equ_B2}
\nonumber  & P\Big\{\underset{ g \in \mathcal{G}_{n, j, \rho}}{\sup} \Big[ \E[g(Z_0)] - \dfrac{1}{n} \sum_{i= 1}^n g(Z_i) \Big] > \dfrac{2^j \rho}{2} \Big\} 
\\
\nonumber & \leq  \exp\left(2 \frac{2^j }{\lambda_n}(L_n + 1) \log \left(\frac{4 \mathcal{K}_{\ell}(L_n + 1)(N_n + 1)B_n}{ 2^j \rho } \right)   - \dfrac{n^{1/(\mu + 2)} (2^j \rho)^{1/(\mu + 2)}/16}{ 4C_{2, n}^{1/\mu + 2 } (1/4)^{(2 \mu + 3)/(\mu + 2)}}  \right)
\\
\nonumber & \leq  \exp\left(2 \frac{2^j }{\lambda_n}(L_n + 1) m \left(  \left(\frac{4 \mathcal{K}_{\ell}(L_n + 1)(N_n + 1)B_n}{ 2^j \rho } \right)^{1/m} - 1  \right)  - \dfrac{n^{1/(\mu + 2)} (2^j \rho)^{1/(\mu + 2)}/16}{ 4C_{2, n}^{1/\mu + 2 } (1/4)^{(2 \mu + 3)/(\mu + 2)}}  \right)
\\
& \leq  \exp\left( 2 \frac{(2^j \rho)^{1 - 1/m} }{\lambda_n}(L_n + 1) m  \left(4 \mathcal{K}_{\ell}(L_n + 1)(N_n + 1)B_n  \right)^{1/m}  - \dfrac{n^{1/(\mu + 2)} (2^j \rho)^{1/(\mu + 2)}/16}{ 4C_{2, n}^{1/\mu + 2 } (1/4)^{(2 \mu + 3)/(\mu + 2)}}  \right).
\end{align}
Let $ 1 - \dfrac{1}{m} = \dfrac{1}{\mu + 2} $. We have
\begin{align}\label{equ_B2}
\nonumber  & P\Big\{\underset{ g \in \mathcal{G}_{n, j, \rho}}{\sup} \Big[ \E[g(Z_0)] - \dfrac{1}{n} \sum_{i= 1}^n g(Z_i) \Big] > \dfrac{2^j \rho}{2} \Big\} 
\\
& \leq  \exp\left(  (2^j \rho)^{1/(\mu + 2)} \left( \frac{ 2 (L_n + 1) (\mu  + 2 )}{\lambda_n (\mu + 1)}  \left(4 \mathcal{K}_{\ell}(L_n + 1)(N_n + 1)B_n \right)^{(\mu + 1)/(\mu + 2)}  - \dfrac{n^{1/(\mu + 2)} /16}{ 4C_{2, n}^{1/\mu + 2 } (1/4)^{(2 \mu + 3)/(\mu + 2)}} \right) \right).
\end{align}
Let $\nu_4 + \nu_1 + \nu_2  < 1/(\mu + 2)$. 
Under the conditions on (\ref{condi_of_seq}) above, there exists $ N_2 \in \N$ such the if $ n \ge N_2$ we have
\[ \frac{ 2 (L_n + 1) (\mu  + 2 )}{\lambda_n (\mu + 1)}  4 \mathcal{K}_{\ell}(L_n + 1)(N_n + 1)B_n    \leq  \dfrac{n^{1/(\mu + 2)} /16}{ 8C_{2, n}^{1/\mu + 2 } (1/4)^{(2 \mu + 3)/(\mu + 2)}}.      \]
Thus
\begin{align}\label{equ_B2}
P\Big\{\underset{ g \in \mathcal{G}_{n, j, \rho}}{\sup} \Big[ \E[g(Z_0)] - \dfrac{1}{n} \sum_{i= 1}^n g(Z_i) \Big] > \dfrac{2^j \rho}{2} \Big\} \leq
& \exp \Bigg(- \left(2^{1/(\mu + 2)}\right)^j \dfrac{n^{1/(\mu + 2)} /16}{ 8C_{2, n}^{1/\mu + 2 } (1/4)^{(2 \mu + 3)/(\mu + 2)}}  
 \rho^{1/(\mu + 2)} \Bigg). 
\end{align}
Let
\[ \beta_n \coloneqq   \dfrac{n^{1/(\mu + 2)} /16}{ 8C_{2, n}^{1/\mu + 2 } (1/4)^{(2 \mu + 3)/(\mu + 2)}}  
 \rho^{1/(\mu + 2)}.     \]
Using similar argument as  (\ref{boundbetan}) we have
\begin{equation}
P( B_{1, n} > \rho ) \leq 2 \exp\left( -\dfrac{n^{1/(\mu + 2)} /16}{ 8C_{2, n}^{1/\mu + 2 } (1/4)^{(2 \mu + 3)/(\mu + 2)}}  
 \rho^{1/(\mu + 2)} \right).
\end{equation}
Hence using similar argument as \textbf{Step 1} we have

\begin{align}
\nonumber \displaystyle\int_{\rho_n}^{1} P( B_{1, n} > \rho ) d \rho  & \leq 2 \displaystyle\int_{\rho_n}^{1} \exp\left( -\dfrac{n^{1/(\mu + 2)} /16}{ 8C_{2, n}^{1/\mu + 2 } (1/4)^{(2 \mu + 3)/(\mu + 2)}}  
 \rho^{1/(\mu + 2)} \right) d \rho
 \\
\nonumber & \leq 2 \dfrac{(128)^{(2(\mu + 2))} (1/4)^{(2(2 \mu + 3))} C_{2, n}^2 }{n^2} \displaystyle\int_{\rho_n}^{1}   \dfrac{1}{\rho^2}d \rho
\\
\nonumber & \leq \dfrac{2 (128)^{(2(\mu + 2))} (1/4)^{(2(2 \mu + 3))} C_{2, n}^2 }{\left( C_{1, n}/(2 C_{2, n}^{1/(\mu + 2)})\right)^{(\mu + 2)/(2 \mu + 3)}n^{(3\mu + 5)/(2\mu + 3)}}  -2 \dfrac{(128)^{(2(\mu + 2))} (1/4)^{(2(2 \mu + 3))} C_{2, n}^2 }{n^2}
\\
& \leq   \dfrac{ (128)^{(2(\mu + 2))} (1/4)^{(2(2 \mu + 3))}C_{2, n}^2 }{\left( C_{1, n}/(2 C_{2, n}^{1/(\mu + 2)})\right)^{(\mu + 2)/(2 \mu + 3)}n^{(3\mu + 5)/(2\mu + 3)}}.
\end{align}

Let,
\begin{align}\label{equarhau}
\nonumber 2 C_{2, n}^{1/\mu + 2 } (n 2^j \rho/4)^{(2 \mu + 3)/(\mu + 2)} & < C_{1, n} n 
\\
\Rightarrow \rho & <  \dfrac{\left( C_{1, n}/(2 C_{2, n}^{1/(\mu + 2)})\right)^{(\mu + 2)/(2 \mu + 3)}}{n^{\mu + 1 / (2\mu + 3)}} \coloneqq \rho_n.
\end{align}
One can see from (\ref{equarhau}) that 
\[- \dfrac{n^2 (2^j \rho)^2/16}{ C_{1, n} n + 2 C_{2, n}^{1/\mu + 2 } (n 2^j \rho/4)^{(2 \mu + 3)/(\mu + 2)} } \leq - \dfrac{n^2 (2^j \rho)^2/16}{ 2 C_{1, n}}.   \]

\textbf{Step 3}: $ 0 < \rho   < \rho_n  $.
\begin{align}
\nonumber \displaystyle\int_{0}^{\rho_n} P( B_{1, n} > \rho ) d \rho  & \leq  \displaystyle\int_{0}^{\rho_n} d \rho \leq \rho_n = \dfrac{\left( C_{1, n}/(2 C_{2, n}^{1/(\mu + 2)})\right)^{(\mu + 2)/(2 \mu + 3)}}{n^{\mu + 1 / (2\mu + 3)}}.
\end{align}
Hence  set  $ N = \max (N_1, N_2)$ which implies
\begin{align}
\nonumber \E [B_{1, n}] \leq \int_{0}^{\infty} P( B_{1, n} > \rho ) d 
\rho & \leq  \dfrac{\left( C_{1, n}/(2 C_{2, n}^{1/(\mu + 2)})\right)^{(\mu + 2)/(2 \mu + 3)}}{n^{\mu + 1 / (2\mu + 3)}} + \dfrac{ (128)^{2(\mu + 2)} C_{2, n}^2 }{\left( C_{1, n}/(2 C_{2, n}^{1/(\mu + 2)})\right)^{(\mu + 2)/(2 \mu + 3)}n^{3\mu + 5/(2\mu + 3)}}  \\
& \hspace{8cm} +  2 \dfrac{(128)^{2(\mu + 2)} C_{2, n}^2 }{n^2}.
\end{align}
For bounded $B_{2, n}$ let us consider $h_{n}^{\circ} \in \mathcal{H}_{\sigma} (L_n, N_n, B_n, F) $ such that  
\begin{equation}
R (h_{n}^{\circ})  + J_{\lambda_n, \tau_n} (h_{n}^{\circ}) \leq  \underset{h \in \mathcal{H}_{\sigma, n}} {\inf} \left[ R(h)  + J_{\lambda_n, \tau_n} (h) \right] + \dfrac{1}{n}.
\end{equation}
Recall the inequality:
\begin{equation}
\widehat{R}_n (\widehat{h}_n) + J_{\lambda_n, \tau_n} ( \widehat{h}_n) \leq  \widehat{R}_n (h) + J_{\lambda_n, \tau_n} (h), ~ \text{for all} ~ h \in \mathcal{H}_{\sigma} (L_n, N_n, B_n, F).
\end{equation}
Thus 
\begin{align} 
\nonumber B_{2, n} & = 2[\widehat{R}_n (\widehat{h}_n) - \widehat{R}_n (h^{*}) ] + 2 J_{\lambda_n, \tau_n} (\widehat{h}_n)
\\
\nonumber & = 2[\widehat{R}_n (\widehat{h}_n) + J_{\lambda_n, \tau_n} (\widehat{h}_n) - \widehat{R}_n (h_{n}^{\circ}) ] 
 + 2 [ \widehat{R}_n (h_{n}^{\circ}) -  \widehat{R}_n (h^{*}) ]
 \\
 \nonumber &  \leq 2 J_{\lambda_n, \tau_n} (h_{n}^{\circ}) + 2 [ \widehat{R}_n (h_{n}^{\circ}) -  \widehat{R}_n (h^{*}) ].
\end{align}
Thus,
\begin{equation}\label{B_2_n_inf}
\E t[  B_{2, n} t] \leq  2 J_{\lambda_n, \tau_n} (h_{n}^{\circ}) + 2 \mathcal{E}_{Z_0} (h_{n}^{\circ}) \leq  2  \underset{h \in \mathcal{H}_{\sigma} (L_n, N_n, B_n, F)} {\inf} \Big[ \mathcal{E}_{Z_0} (h)  + J_{\lambda_n, \tau_n} (h) \Big] + \dfrac{1}{n}.
\end{equation}
Where $ \mathcal{E}_{Z_0} (h) =  R(h) -  R (h^{*}) $.
Thus, 
\begin{align}
\nonumber \mathcal{E}_{Z_0} (\widehat{h}_n) 
& \leq 2  \Big\{ \underset{h \in \mathcal{H}_{\sigma} (L_n, N_n, B_n, F)} {\inf} \Big[ \mathcal{E}_{Z_0} (h)  + J_{\lambda_n, \tau_n} (h) \Big]  \Big\}\\
& \nonumber \lor 
\Bigg( \dfrac{\left( C_{1, n}/(2 C_{2, n}^{1/(\mu + 2)})\right)^{(\mu + 2)/(2 \mu + 3)}}{n^{\mu + 1 / (2\mu + 3)}} + 
 \dfrac{ (128)^{2(\mu + 2)} C_{2, n}^2 }{\left( C_{1, n}/(2 C_{2, n}, n^{1/(\mu + 2)})\right)^{(\mu + 2)/(2 \mu + 3)}n^{3\mu + 5/(2\mu + 3)}}  
 +  2 \dfrac{(128)^{2(\mu + 2)} C_{2, n}^2 }{n^2} \Bigg)
\\
& \nonumber \lesssim  2  \Big\{ \underset{h \in \mathcal{H}_{\sigma, n}} {\inf} \Big[ \mathcal{E}_{Z_0} (h)  + J_{\lambda_n, \tau_n} (h) \Big]  \Big\} \lor \dfrac{C_{\mu} \left( C_{1, n}/(2 C_{2, n}^{1/(\mu + 2)})\right)^{(\mu + 2)/(2 \mu + 3)}}{n^{\mu + 1 / (2\mu + 3)}},
\end{align}
with a positive  constant $ C_{\mu}$ depending of $\mu$. Which completes the proof  of the Theorem.
\qed

\subsection{Proof of Theorem \ref{thm_3}}
Let us consider the following decomposition:
\[ R(\widehat{h}_n) - R(h^{*})   \coloneqq B_{1, n} +  B_{2, n}. \]
Where,
\[
\begin{array}{llll}
  B_{1, n}   & =   [R(\widehat{h}_n) - R(h^{*})]  - 2 [ \widehat{R}_n (\widehat{h}_n)  - \widehat{R}_n (h^{*}) ] - 2 J_{\lambda_n, \tau_n} (\widehat{h}_n); 
  \\
 B_{2, n}   &  =   2[\widehat{R}_n (\widehat{h}_n) - \widehat{R}_n (h^{*})] + 2 J_{\lambda_n, \tau_n} (\widehat{h}_n).
\end{array}
\]
For bound $B_{1, n}$, we use similar arguments of the proof of Theorem \ref{thm4}.

 Let $l = \mathcal{N}( \varepsilon, \mathcal{G}_ {n, j, \rho}, \| \cdot \|_\infty) $. We have from (\ref{equa2}),
\begin{align}\label{equ_br}
 P \Big\{\underset{ g \in \mathcal{G}_{n, j, \rho}}{\sup} \Big[ \E[g(Z_0)] - \dfrac{1}{n} \sum_{i= 1}^n g(Z_i) \Big] > \varepsilon \Big\}  & \leq \mathcal{N} \left(\varepsilon, \mathcal{G}_{n, j, \rho} \right) \exp \Big ( - n^{\nu_5} \varepsilon +  \dfrac{  n^{2 \nu_5 - 1} (M + \theta_{\infty, n} (1))^2}{2} \Big).
\end{align}
 In \cite{ohn2022nonconvex}, we have  
\begin{equation}\label{P_inqu2}
\mathcal{N}(\varepsilon, \mathcal{G}_{n, j, \rho}, \| \cdot \|_\infty) \leq \mathcal{N} \left(\frac{\varepsilon}{\mathcal{K_{\ell}}}, \mathcal{H}_{n, j, \rho}, \| \cdot \|_\infty \right).
\end{equation} 
One can easily see that,
\begin{equation}\label{inclusion}
 \mathcal{H}_{n, j, \rho}  \subset \left\{  h \in \mathcal{H}_{\sigma}(L_{n}, N_{n}, B_{n}, F, \dfrac{2^j \rho}{\lambda_n}): \| \theta(h) \|_{ \text{clip}, \tau_n} \leq \frac{2^j \rho}{\lambda_n}  \right\}. 
 \end{equation}
Let $ \Delta (h_1), ~  \Delta (h_2) \in \mathcal{G}_{n, j, \rho}, ~ \text{and} ~ (x, y) \in \R^d \times \R$, we have
\begin{align}
\nonumber \| \Delta (h_1) (x, y) -  \Delta (h_2) (x, y) \|_{\infty} & = \left| \ell (h_1 (x), y) - \ell (h_2 (x), y) \right| 
\\
&  \leq  \mathcal{K}_{\ell} \left| h_1 (x) - h_2 (x) \right|  \quad \text{from} ~ \textbf{(A3)}.
\end{align}
Thus, in \cite{ohn2022nonconvex}, we have the following inequality
\begin{align}\label{cover_number}
\mathcal{N}  ( \varepsilon, \mathcal{G}_{n, j, \rho}, \| \cdot \|_\infty )  \nonumber & \leq \mathcal{N} \left( \dfrac{\varepsilon}{\mathcal{K}_{\ell}}, \mathcal{H}_{n, j, \rho}, \| \cdot \|_\infty \right) 
\\
\nonumber & \leq \mathcal{N} \left( \frac{\varepsilon}{\mathcal{K}_{\ell}}, \mathcal{H}_{\sigma}(L_{n},N_{n}, B_{n}, F, \frac{2^j \rho}{\lambda_n}),  \| \cdot \|_\infty \right)
\\
& \leq  \exp\left( 2 \frac{2^j \rho}{\lambda_n}(L_n + 1) \log \left(\frac{(L_n + 1)(N_n + 1)B_n}{\dfrac{\varepsilon}{\mathcal{K}_{\ell}} - \tau_n (L_n + 1)((N_n + 1) B_n)^{L_n +1}} \right) \right).
\end{align} 
Thus 
\begin{align}\label{equ_B1}
\nonumber  & P\Big\{\underset{ g \in \mathcal{G}_{n, j, \rho}}{\sup} \Big[ \E[g(Z_0)] - \dfrac{1}{n} \sum_{i= 1}^n g(Z_i) \Big] > \varepsilon \Big\} 
\\
\nonumber & \leq  \exp\left(2 \frac{2^j \rho}{\lambda_n}(L_n + 1) \log \left(\frac{(L_n + 1)(N_n + 1)B_n}{\dfrac{\varepsilon}{\mathcal{K}_{\ell}} - \tau_n (L_n + 1)((N_n + 1) B_n)^{L_n +1}} \right) \right)  \exp \Bigg ( - n^{\nu} \varepsilon +  \dfrac{  n^{2 \nu - 1} (M + \theta_{\infty, n} (1))^2}{2} \Bigg)
\\
& \leq   \exp\left(2 \frac{2^j \rho}{\lambda_n}(L_n + 1) \log \left(\frac{(L_n + 1)(N_n + 1)B_n}{\dfrac{\varepsilon}{\mathcal{K}_{\ell}} - \tau_n (L_n + 1)((N_n + 1) B_n)^{L_n +1}} \right) - n^{\nu_5} \varepsilon +  \dfrac{  n^{2 \nu_5 - 1} (M + \theta_{\infty, n} (1))^2}{2} \right). 
\end{align}
Let $ \varepsilon = \dfrac{2^j \rho}{2}$,  we have
\begin{align}\label{equ_B2}
\nonumber  & P\Big\{\underset{ g \in \mathcal{G}_{n, j, \rho}}{\sup} \Big[ \E[g(Z_0)] - \dfrac{1}{n} \sum_{i= 1}^n g(Z_i) \Big] > \dfrac{2^j \rho}{2} \Big\} 
\\
 & \leq  \exp\left(2 \frac{2^j \rho}{\lambda_n}(L_n + 1) \log \left(\frac{(L_n + 1)(N_n + 1)B_n}{\dfrac{2^j  \rho}{2\mathcal{K}_{\ell}} - \tau_n (L_n + 1)((N_n + 1) B_n)^{L_n +1}} \right)  -  \dfrac{2^j \rho}{2} n^{\nu_5} +  \dfrac{  n^{2 \nu_5 - 1} (M + \theta_{\infty, n} (1))^2}{2}\right).
\end{align}
\textbf{Step 1}: $\rho >  1$.

\noindent Recall the assumption
\[ \tau_n \leq \dfrac{1}{4 \mathcal{K}_{\ell} (L_n + 1)((N_n + 1)B_n)^{L_n + 1}}. \]
Since $ 2^{j} \rho > 1 $ we have 
\begin{align}\label{equ_B3}
\nonumber  & P\Big\{\underset{ g \in \mathcal{G}_{n, j, \rho}}{\sup} \Big[ \E[g(Z_0)] - \dfrac{1}{n} \sum_{i= 1}^n g(Z_i) \Big] > \dfrac{2^j \rho}{2} \Big\} 
\\
 & \leq  \exp\left(2 \frac{2^j \rho}{\lambda_n}(L_n + 1) \log \left( 4 \mathcal{K}_{\ell}(L_n + 1)(N_n + 1)B_n \right)  -  \dfrac{2^j \rho}{2} n^{\nu_5} +  2^j \rho \dfrac{  n^{2 \nu_5 - 1} (M + \theta_{\infty, n} (1))^2}{2}\right)
 \\
 & \leq  \exp\left(2^j \rho \left( \frac{2}{\lambda_n}(L_n + 1) \log \left(4 \mathcal{K}_{\ell}(L_n + 1)(N_n + 1)B_n \right)  + \dfrac{  n^{2 \nu_5 - 1} (M + \theta_{\infty, n} (1))^2}{2} -   \dfrac{n^{\nu_5}}{2}   \right) \right).
 \end{align}
Recall the condition
\begin{equation}\label{condi_sequ_2}
L_n \lesssim \log n, ~ N_n \lesssim n^{\nu_1}, B_n \lesssim n^{\nu_2}, ~ \lambda_n \asymp (\log n)^{\nu_3}/n^{\nu_4 }, ~ \nu_1 + \nu_2 + \nu_4 < \nu_5 \quad \text{for some} ~ \nu_1, ~  \nu_2, ~ \nu_3, ~  \nu_4 > 0, ~ \nu_5 < 1.   
\end{equation}
Under the conditions (\ref{condi_sequ_2}) above, there exists $ N_3 \in \N$ such that if $ n \ge N_3$ we have
\[ \delta_n \coloneqq \frac{2}{\lambda_n}(L_n + 1) \log \left(4 \mathcal{K}_{\ell}(L_n + 1)(N_n + 1)B_n \right)  + \dfrac{  n^{2 \nu_5 - 1} (M + \theta_{\infty, n} (1))^2}{2} <  \dfrac{n^{\nu_5}}{4}.   \]
Thus,  
\begin{align}\label{equ_B3}
 P\Big\{\underset{ g \in \mathcal{G}_{n, j, \rho}}{\sup} \Big[ \E[g(Z_0)] - \dfrac{1}{n} \sum_{i= 1}^n g(Z_i) \Big] > \dfrac{2^j \rho}{2} \Big\} 
 & \leq  \exp\left(-2^j \dfrac{n^{\nu_5}}{4} \rho \right).
 \end{align}
Let $ \beta_n  \coloneqq \dfrac{n^{\nu_5}}{4} \rho $. Using the following inequality $\exp \left( - \beta_n 2^j \right) \leq \exp \left( - \beta_n  \right) 2^{-\frac{\beta_n j}{\log 2}}$,
we get,
\begin{align}\label{boundbeta}
\sum_{j = 1}^n P\Big\{\underset{ g \in \mathcal{G}_{n, j, \rho}}{\sup} \Big[ \E[g(Z_0)] - \dfrac{1}{n} \sum_{i= 1}^n g(Z_i) \Big] > \dfrac{2^j \rho}{2} \Big\}  \leq   \exp \left( - \beta_n  \right) \sum_{j =1}^n  \left( 2^{-\frac{\beta_n }{\log 2}} \right)^j \leq 2 \exp \left( - \beta_n  \right).
\end{align}
By applying (\ref{boundbeta}) we have
\begin{equation}
P \left ( B_{1, n} > \rho \right) \leq 2 \exp \left( - \frac{n^{\nu_5}}{4} \rho \right).
\end{equation}
We have,
\begin{equation}
\displaystyle\int_{1}^{\infty} P \left ( B_{1, n} > \rho \right) d \rho \leq  2 \displaystyle\int_{1}^{\infty} \exp \left( - \frac{n^{\nu_5}}{4} \rho \right) d \rho =  \dfrac{8}{n^{\nu_5}}\exp\left (- \dfrac{n^{\nu_5}}{4}\right).
\end{equation}

\textbf{Step 2}: $  \rho_n < \rho  < 1$.
Recall the assumption
\[ \tau_n \leq \dfrac{\rho_n}{4 \mathcal{K}_{\ell} (L_n + 1)((N_n + 1)B_n)^{L_n + 1}}, ~ \text{with}  ~  \rho_n \coloneqq \dfrac{1}{n^{2 \nu_6}}. \]

\noindent Since $ \rho \in (\rho_n, 1)$ we have 
\begin{align}\label{equ_B4}
\nonumber  & P\Big\{\underset{ g \in \mathcal{G}_{n, j, \rho}}{\sup} \Big[ \E[g(Z_0)] - \dfrac{1}{n} \sum_{i= 1}^n g(Z_i) \Big] > \dfrac{2^j \rho}{2} \Big\} 
\\
\nonumber  & \leq  \exp\left(2 \frac{2^j \rho}{\lambda_n}(L_n + 1) \log \left(\frac{(L_n + 1)(N_n + 1)B_n}{\dfrac{2^j  \rho}{2\mathcal{K}_{\ell}} - \tau_n (L_n + 1)((N_n + 1) B_n)^{L_n +1}} \right)  -  \dfrac{2^j \rho}{2} n^{\nu_5} +  \dfrac{  n^{2 \nu_5 - 1} (M + \theta_{\infty, n} (1))^2}{2}\right)
\\
\nonumber  & \leq  \exp\left(2 \frac{2^j \rho}{\lambda_n}(L_n + 1) \log \left(4 \mathcal{K}_{\ell} \frac{(L_n + 1)(N_n + 1)B_n}{\rho_n}  \right)  -  \dfrac{2^j \rho}{2} n^{\nu_5} +  \dfrac{ 2^{j} \rho n^{2 \nu_5 - 1} (M + \theta_{\infty, n} (1))^2}{2 \rho_n }\right)
\\
 & \leq  \exp\left( 2^j \rho \left ( \frac{2}{\lambda_n}(L_n + 1) \log \left(\frac{4 \mathcal{K}_{\ell}(L_n + 1)(N_n + 1)B_n}{1/n^{2 \nu_6}} \right)  + \dfrac{n^{2\nu_5 + 2\nu_6 -1}(M + \theta_{\infty, n} (1))^{2}}{2} -   \dfrac{n^{\nu_5}}{2}    \right) \right).
\end{align}
Recall the condition
\begin{equation}\label{condi_sequ_3}
L_n \lesssim \log n, ~ N_n \lesssim n^{\nu_1}, B_n \lesssim n^{\nu_2}, ~ \lambda_n \asymp (\log n)^{\nu_3}/n^{\nu_4 }, ~  \nu_4 + 2\nu_6 +
 \nu_1 + \nu_2 < \nu_5 \quad \text{for} ~ \nu_1, ~  \nu_2, ~ \nu_3, ~  \nu_4 > 0, ~ \nu_5 < 1, ~ \nu_6 < \dfrac{1 - \nu_5}{2}.
\end{equation}
Under the conditions  (\ref{condi_sequ_3}) above, there exists $ N_4 \in \N$ such that if $ n \ge N_4$ we have
\[ \frac{2}{\lambda_n}(L_n + 1) \log \left(\frac{4 \mathcal{K}_{\ell}(L_n + 1)(N_n + 1)B_n}{1/n^{2 \nu_6}} \right)  + \dfrac{n^{2\nu_5 + 2\nu_6 -1}(M + \theta_{\infty, n} (1))^{2}}{2} <  \dfrac{n^{\nu_5}}{4}.    \]
Thus,
\begin{align}
\nonumber  & P\Big\{\underset{ g \in \mathcal{G}_{n, j, \rho}}{\sup} \Big[ \E[g(Z_0)] - \dfrac{1}{n} \sum_{i= 1}^n g(Z_i) \Big] > \dfrac{2^j \rho}{2} \Big\} \leq \exp \left( - 2^j \dfrac{n^{\nu_5}}{4} \rho \right)
\end{align}
Let $ \beta_{n,2}  \coloneqq \dfrac{n^{\nu_5}}{4} \rho $.
Thus,
\begin{equation}
P \left( B_{1, n} > \alpha \right) \leq  2 \exp \left( -  \dfrac{n^{\nu_5}}{4}  \rho\right)
\end{equation}
Hence,
\begin{align}
\nonumber \displaystyle\int_{\rho_n}^{1} P \left ( B_{1, n}   > \rho \right) d \rho & \leq 2 \displaystyle\int_{\rho_n}^{1} \exp \left( - \dfrac{n^{\nu_5}}{4} \rho \right) d \rho 
\\
  & \leq  \dfrac{8}{n^{\nu_5}} \exp \left( -  \dfrac{n^{\nu_5 - 2 \nu_6} }{4 } \right) - \dfrac{8}{n^{\nu_5}}\exp \left( -  \dfrac{n^{\nu_5}}{4} \right) \leq  \dfrac{4}{n^{\nu_5}} \exp \left( -  \dfrac{n^{\nu_5 - 2 \nu_6} }{4 } \right).
\end{align}

\medskip

\noindent \textbf{Step 3}: $ 0 \leq \rho \leq \rho_n$.

\noindent We have
\begin{equation}
\displaystyle\int_{0}^{\rho_n} P \left ( B_{1, n} > \rho \right) d \rho \leq \displaystyle\int_{0}^{\rho_n} d \rho = \rho_n= \dfrac{1}{n^{2 \nu_6}}.
\end{equation}
Thus, set $ N = \max (N_3, N_4) $  which implies 
\begin{align}
\nonumber \E [ B_{1, n}]  & \leq \displaystyle\int_{0}^{\infty} P \left ( B_{1, n} > \rho \right) d \rho 
\\
 & \leq  \dfrac{8}{n^{\nu_5}}\exp\left (- \dfrac{n^{\nu_5}}{8}\right) + \dfrac{1}{n^{2 \nu_6}} +  \dfrac{4}{n^{\nu_5}} \exp \left( -  \dfrac{n^{\nu_5 - 2 \nu_6} }{4 } \right).
\end{align}

\medskip

One can bound $B_{2, n}$ as in the proof of Theorem \ref{thm4}, see also (\ref{B_2_n_inf}).
%
%
%

Thus,
\begin{align}
\nonumber \mathcal{E}_{Z_0} (\widehat{h}_n) & =  R(\widehat{h}_n) -  R (h^{*}) 
\\
\nonumber & \leq 2  \Big\{ \underset{h \in \mathcal{H}_{\sigma, n}} {\inf} \Big[ \mathcal{E}_{Z_0} (h)  + J_{\lambda_n, \tau_n} (h) \Big]  \Big\} \lor \Bigg(  \dfrac{8}{n^{\nu_5}}\exp\left (- \dfrac{n^{\nu_5}}{8}\right) + \dfrac{1}{n^{2 \nu_6}} +  
  \dfrac{4}{n^{\nu_5}} \exp \left( -  \dfrac{n^{\nu_5 - 2 \nu_6} }{4 } \right)
\\
& \lesssim 2  \Big\{ \underset{h \in \mathcal{H}_{\sigma, n}} {\inf} \Big[ \mathcal{E}_{Z_0} (h)  + J_{\lambda_n, \tau_n} (h) \Big]  \Big\} \lor \dfrac{C}{n^{2 \nu_6}},
\end{align}
for a  positive constant $C$.
Hence, the Theorem follows. \qed

\subsection{Proof of Corollary \ref{corol1}}
Let $L_n, ~ N_n, ~ S_n,~  B_n, ~ F_n > 0$ and consider the class of DNNs $ \mathcal{H}_{\sigma, n} = \mathcal{H}_{\sigma} (L_n, N_n, S_n, B_n, F_n ) \subset \mathcal{F} (\mathcal{X}, \mathcal{Y}) $.
Under the Lipschitz property of $\ell$, for all $ h \in \mathcal{H}_{\sigma, n}$, it holds that,
\begin{equation}
R(h) - R(h^{*}) = \E_{Z_0} [\ell(h(X_0), Y_0)] - \E_{Z_0} [\ell(h^{*}(X_0), Y_0)] \leq \mathcal{K}_{\ell} \E_{X_0} [ |h(X_0) - h^{*} (X_0) |].
\end{equation}
Since $ h^{*} \in C^{s, \mathcal{K}} (\mathcal{X})$ for some $\mathcal{K} > 0$, then, for all $  \epsilon > 0$, we can find a positive constants $ L_0, N_0, S_0, B_0$ such that, with
\begin{equation}
L_n = L_0 \log_{+}(1/\epsilon), ~ N_n = N_0 \epsilon^{-d/s}, ~ S_n = S_0 \epsilon^{-d/s} \log_{+} (1/\epsilon) ~ \text{and} ~ B_n = B_0 \epsilon^{-4(d/(s + 1))},
\end{equation}
there is a neural network $h_{\epsilon} \in \mathcal{H}_{\sigma, n}$ satisfying,
\begin{equation}\label{cond_excess_risk_bound}
\| h_{\epsilon} - h^{*} \|_{\infty, \mathcal{X}} < \epsilon,
\end{equation}
see \cite{kengne2023deep}; and where $\log_{+} (x) = \max (1, \log x)$ for all $ x > 0$.
Hence,
\[ \E_{X_0} \left[ | h_{\epsilon} (X_0) - h^{*}(X_0) |\right] < \epsilon. 
\]
Let $ \epsilon_n = n^{-\dfrac{\nu_4}{(1 + d/s)}}$. One can easily see that, the conditions $ L_n \lesssim \log n, ~ N_n \lesssim n^{\nu_1}, ~ 1\leq B_n \lesssim n^{\nu_2}$ are satisfied with $ \nu_1 = \dfrac{d \nu_4}{s(1 + d/s)}, ~ \nu_2 = \dfrac{4 d \nu_4}{(s + 1)(1 + d/s)} $. 
Thus, by using Theorem (\ref{thm_3}) and (\ref{cond_excess_risk_bound}), and the fact that for any $h \in \mathcal{H}_{\sigma, n}$,
 $ \|\theta (h)\|_{\text{clip}, \tau} \leq \| \theta(h)\|_0$ for any $\tau > 0$, we have,
\begin{align}
\nonumber \E  [\mathcal{E}_{Z_0} (\widehat{h}_n) ] & \lesssim 2  \Big\{ \underset{h \in \mathcal{H}_{\sigma} (L_n, N_n, S_n, B_n, F)} {\inf} \Big[ \mathcal{E}_{Z_0} (h)  + J_{\lambda_n, \tau_n} (h) \Big]  \Big\} \lor \dfrac{C}{n^{2 \nu_6}}
\\
\nonumber &  \lesssim 2  \Big\{ \underset{h \in \mathcal{H}_{\sigma} (L_n, N_n, S_n, B_n, F)} {\inf} \Big[\mathcal{E}_{Z_0} (h) + \lambda_n S_0 \epsilon_n^{-d/s} \log (1/\epsilon_n) \Big] \Big\}  \lor \dfrac{C}{n^{2 \nu_6}}
\\
\nonumber &  \lesssim 2  \Big\{ \underset{h \in \mathcal{H}_{\sigma} (L_n, N_n, S_n, B_n, F)} {\inf} \left[\mathcal{E}_{Z_0} (h) \right] + \lambda_n S_0 \epsilon_n^{-d/s} \log (1/\epsilon_n) \Big\}  \lor \dfrac{C}{n^{2 \nu_6}}
\\
\nonumber &  \lesssim 2  \Big\{ \mathcal{K}_{\ell} \E_{X_0} \left[ | h_{\epsilon_n} (X_0) - h^{*} (X_0) |\right] + \lambda_n S_0 \epsilon_n^{-d/s} \log (1/\epsilon_n) \Big\}  \lor \dfrac{C}{n^{2 \nu_6}}
\\
\nonumber &  \lesssim 2  \Big ( \mathcal{K}_{\ell} \epsilon_n + \lambda_n S_0 \epsilon_n^{-d/s} \log (1/\epsilon_n) \Big ) \lor \dfrac{C}{n^{2 \nu_6}}
\\
&  \lesssim \dfrac{  2 \nu_4(\log n)^{\nu_3 + 1}}{(1 + d/s) n^{\nu_4/(1 + d/s)}} \lor \dfrac{C}{n^{2 \nu_6}}.
\end{align}

\end{document}